\documentclass[authoryear,12pt,sort&compress,times]{elsarticle}
\usepackage[utf8]{inputenc}
\usepackage{latexsym,amsmath,amscd,amssymb,graphics}
\usepackage{enumerate}
\usepackage{wrapfig}
\usepackage{graphicx}
\usepackage{framed}
\usepackage{color}
\usepackage{float}
\usepackage{url}

\usepackage[all]{xy}

\makeatletter

\newtheorem{theorem}{Theorem}

\newtheorem{definition}[theorem]{Definition}

\newtheorem{remark}[theorem]{Remark}

\numberwithin{theorem}{section}

\def\be{\begin{equation}}
\def\ee{\end{equation}}
\def\bea{\begin{eqnarray}}
\def\eea{\end{eqnarray}}
\def\ba{\begin{array}}
\def\ea{\end{array}}

\def\bOm{\boldsymbol{\Omega}}
\def\bp{\mathbf{p}}
\def\bx{\mathbf{x}}

\newcommand{\rem}[1]{}

\newcommand{\de}{\delta}

\newcommand{\bq}{\mathbf{q}}
\newcommand{\bu}{\mathbf{u}}

\newcommand{\bv}{\boldsymbol{v}}

\newcommand{\bxi}{\boldsymbol{\xi}}

\newcommand{\bPi}{\boldsymbol{\Pi}}

\newcommand{\bPhi}{\boldsymbol{\Phi}}

\newcommand{\pp}[2]{\frac{\partial #1}{\partial #2}}

\newcommand{\mse}{\mathfrak{se}}

\newcommand{\todo}[1]{\vspace{5 mm}\par \noindent
\framebox{\begin{minipage}[c]{0.95 \textwidth}
\tt #1 \end{minipage}}\vspace{5 mm}\par}
\begin{document}

\begin{frontmatter}

\title{Lie-Poisson Neural Networks (LPNets): Data-Based Computing of Hamiltonian Systems with Symmetries}
\author[1]{Christopher Eldred}
\ead{celdred@sandia.gov}
\author[2]{Fran\c{c}ois Gay-Balmaz}
\ead{francois.gay-balmaz@lmd.ens.fr}
\author[3]{Sofiia Huraka}
\ead{sophiahuraka@gmail.com}
\author[3]{Vakhtang Putkaradze\corref{cor1}}
\cortext[cor1]{Corresponding author}
\ead{putkarad@ualberta.ca}

\affiliation[1]{organization = {Computer Science Research Institute}, 
addressline={Sandia National Laboratory, 1450 Innovation Pkwy SE}, city={Albuquerque}, state = {NM}, postcode={87123}, country={USA}}

\affiliation[2]{organization = {LMD - Ecole Normale Superieure de Paris}, adderessline={CNRS}, postcode={75005}, city={Paris}, country={France}} 

\affiliation[3]{organization={Department of Mathematical and Statistical Sciences},
            addressline={University of Alberta}, 
            city={Edmonton},
            postcode={T6G 2G1}, 
            state={Alberta},
            country={Canada}}

%\affiliation[1]{LMD - Ecole Normale Sup\'erieure de Paris - CNRS, 75005, Paris; email: gaybalma@lmd.ens.fr} 

\date{\today}
%\maketitle

\begin{abstract} 
An accurate data-based prediction of the long-term evolution of Hamiltonian systems requires a network that preserves the appropriate structure under each time step.  Every Hamiltonian system contains two essential ingredients: the Poisson bracket and the Hamiltonian. Hamiltonian systems with symmetries, whose paradigm examples are the Lie-Poisson systems, have been shown to describe a broad category of physical phenomena, from satellite motion to underwater vehicles, fluids, geophysical applications, complex fluids, and plasma physics. The Poisson bracket in these systems comes from the symmetries, while the Hamiltonian comes from the underlying physics. We view the symmetry of the system as primary, hence the Lie-Poisson bracket is known exactly, whereas the Hamiltonian is regarded as coming from physics and is considered not known, or known approximately. Using this approach, we develop a network based on transformations that exactly preserve the Poisson bracket and the special functions of the Lie-Poisson systems (Casimirs) to machine precision. We present two flavors of such systems: one, where the parameters of transformations are computed from data using a dense neural network (LPNets), and another, where the composition of transformations is used as building blocks (G-LPNets).  We also show how to adapt these methods to a larger class of Poisson brackets. We apply the resulting methods to several examples, such as rigid body (satellite) motion, underwater vehicles, a particle in a magnetic field, and others. The methods developed in this paper are important for the construction of accurate data-based methods for simulating the long-term dynamics of physical systems.
\end{abstract}

\end{frontmatter}

\section{Introduction}

\subsection{Relevant work}

Machine Learning (ML) approaches to data assimilation and modeling have been successful in interpreting large amounts of unstructured data. However, for many scientific and industrial purposes, direct applications of ML methods for big data have been developed without understanding that the underlying engineering and physics are challenging. To address these problems, \textit{Physics Informed Neural Networks (PINNs)} have been developed \citep{raissi2019physics}. As it is impossible to provide a complete overview of the PINNs literature here, we refer the reader to recent reviews \citep{karniadakis2021physics,cuomo2022scientific} for references and more thorough review of the literature. In that approach, one assumes that the motion of the system $\bu(t)$ is approximated by a law of motion 
\begin{equation} 
\dot{\bu} =\mathbf{f} (\bu,t), 
\label{general_eq}
\end{equation} 
where $\bu$ and $\mathbf{f} $ can be either finite-dimensional, forming a system of ODEs, or infinite dimentional, forming a system of PDEs. In the 'vanilla' PINNs approach, the initial conditions, boundary conditions, and the ODEs itself form a part of the loss function optimized during the learning procedure. The last part -- computing the difference $\dot{\bu} - \mathbf{f}(\bu,t)$ -- can be achieved for a neural network approximating $\bu$, that is, approximating the mapping $(\bu_0,t) \rightarrow \bu$. 
Similarly, for PDEs in a ($1+1$) dimension, for example, the solution $u(x,t)$ is a mapping from the position $x$ and time $t$ to the space of solutions. In PINNs, one approximates this mapping by a neural network. For a given set of weights and activating functions in neural network, at the given set of points in space and time $(t_i,\bx_i)$, one can compute the \emph{exact} values of temporal and spatial derivatives using the method of automatic differentiation \citep{baydin2018automatic}. One then constructs the Mean-Square-Error (MSE) of the PDE approximation of solution and boundary conditions, taken as the cost function,  and optimizes network weights to minimize that cost. The advantage of PINNs is their computational efficiency: speedups more than 100,00x for evaluations of solutions of complex systems like weather have been reported \citep{bi2023accurate,pathak2022fourcastnet}, although the quantification of speedup still needs to be understood \citep{karniadakis2021physics}. PINNs are thus extremely useful in practical applications, and have been implemented as the open source using \emph{Python} and \emph{Julia}, as well as Nvidia's \emph{Modulus}.

In spite of the successes, there are still many uncertainties in the applications of PINNs. For example, it is known that PINNs may struggle with systems having dynamics with widely different time scales \citep{karniadakis2021physics}, or having trouble in finding the true minimum for optimization \citep{krishnapriyan2021characterizing}. Finally, systems with very little friction, like Hamiltonian systems, are difficult to describe by the standard PINNs, since evaluation errors created on every time step tend to accumulate with no chance of dissipating quickly enough. This paper develops a method of simulating a type of non-canonical Hamiltonian system, namely, Lie-Poisson systems. These systems are extremely important in building models of physical systems in Eulerian or body coordinates.  The method is also shown to generalize to some other types of non-canonical Poisson brackets. The remainder of the literature review is dedicated to  work on the use of data-based methods in simulating Hamiltonian systems. In order to make the discussion more concrete, we introduce a brief background in Hamiltonian and Poisson systems. More details on this can be found in Section \ref{sec:intro_EP}.

There are several approaches applying physics-informed data methods to computations of Hamiltonian systems. Most of the work has been focused on \emph{canonical} Hamiltonian systems, \emph{i.e.}, the systems where the law of motion \eqref{general_eq} has the following particular structure: $\bu$ is $2n$-dimensional,  $\bu=(\bq, \bp)$, and there is a function $H(\bq,\bp)$, the Hamiltonian, such that $\mathbf{f} $ in \eqref{general_eq} becomes 
\begin{equation}
\mathbf{f}  = \mathbb{J} \nabla_{\bu} H \, , \quad  
\mathbb{J} = 
\left( 
\begin{array}{cc}
0 & \mathbb{I}_n  
\\
- \mathbb{I}_n & 0
\end{array}
\right) \,,
\label{canonical_system_gen}
\end{equation}
with $\mathbb{I}_n$ the $n \times n$ identity matrix, 
leading to the \emph{canonical} Hamilton equations for $(\bq, \bp)$: 
\begin{equation}
\dot \bq =  \pp{H}{\bp} \, , \quad 
\dot \bp = - \pp{H}{\bq} \,.
\label{canonical_system}
\end{equation}
The evolution of an arbitrary phase space function $F(\bq,\bp)$ along a solution of \eqref{canonical_system} is then described by the canonical Poisson bracket : 
\begin{equation}
\frac{dF}{dt} = \left\{ F, H \right\} = 
\pp{F}{\bq}\pp{H}{\bp}
- 
\pp{H}{\bq}\pp{F}{\bp}.
\label{canonical_bracket}
\end{equation}
The bracket \eqref{canonical_bracket} is a mapping sending two smooth functions of $(\bq,\bp)$ into a smooth function of the same variables. This mapping is bilinear, antisymmetric, acts as a derivation on both functions, and satisfies the Jacobi identity: for all functions $F,G,H$
\begin{equation}
\{ \{ F, G \}, H \} + \{ \{ H, F \}, G \} + 
\{ \{ G, H \}, F \} =0 \, . 
\label{Jacobi_identity}
\end{equation}
%\todo{VP: My changes in magenta}
%\textcolor{magenta}{A more general type of Hamiltonian system, known as non-canonical or Poisson systems. Often, these brackets have a non-trivial null space leading to the conservation of certain quantities known as the Casimir constants, or simply  \emph{Casimirs}. The Casimirs are properties of a particular bracket and thus are independent of a particular realization of a given Hamiltonian. } 

Brackets that satisfy all the required properties, \emph{i.e.}, are bilinear, antisymmetric, act as a derivation and satisfy Jacobi identity \eqref{Jacobi_identity}, but are not described by the canonical equations \eqref{canonical_bracket}, are called general Poisson brackets (also known as non-canonical Poisson brackets). The corresponding equations of motion are called non-canonical Hamiltonian (or Poisson) systems. Often, these brackets have a non-trivial null space leading to the conservation of certain quantities known as the Casimir constants, or simply  \emph{Casimirs}. The Casimirs are properties of the Poisson bracket and are independent of a particular realization of a given Hamiltonian. This paper will focus on the data-based approaches for computations of an important class of non-canonical Poisson systems. 

There is an avenue of thought that focuses on learning the actual Hamiltonian for the system from data, or, in the more general case, the Poisson bracket. 
This approach was explicitly implemented for canonical Hamiltonian systems in \citep{greydanus2019hamiltonian} under the name of \textit{Hamiltonian Neural Networks (HNN)}, which approximated the Hamiltonian function $H(\bq,\bp)$ fitting evolution of the particular data sequence through equations \eqref{canonical_system}. It was shown that embedding the dynamics with the knowledge of the data allows a much more accurate and robust way to approximate the solution compared to a general Neural Network (NN). This work was further extended to include the adaptive learning of parameters and transitions to chaos \citep{han2021adaptable}. The mathematical background guaranteeing the existence of Hamiltonian function sought in HNN was derived in \citep{david2021symplectic}. 
An alternative method of learning the equations is given by the \textit{Lagrangian Neural Networks (LNNs)} \citep{cranmer2020lagrangian} which approximates the solutions of Euler-Lagrange equations, \emph{i.e.}, the equations in the coordinate-velocity space $(\bq, \dot \bq)$ \emph{before} Legendre-transforming to the momentum-coordinate representation $(\bq, \bp)$ given by equations \eqref{canonical_system}. More generally, learning a vector field for the non-canonical Poisson brackets was suggested in \citep{vsipka2023direct}. The main challenge in that work was enforcing Jacobi identity \eqref{Jacobi_identity} for the learned equation structure. 

In these and other works on the topic, one learns the vector field governing the system, with the assumption that the vector field can be solved using appropriate numerical methods. However, one needs to be aware that care must be taken in computing the numerical solutions for Hamiltonian systems, especially for long-term computations, as regular numerical methods lead to distortion of quantities that should be conserved, such as total energy and, when appropriate, the momenta. 
In order to compute long-term evolution of systems obeying the Hamiltonian vector fields, whether exact or approximated by the Hamiltonians derived from neural networks, one can use variational integrator methods \citep{marsden2001discrete,leok2012general,hall2015spectral} that conserve momenta-like quantities with machine precision. However, these integrators may be substantially more computationally intensive compared to non-structure preserving methods.

In this paper, we focus on an alternative approach, namely, exploring the learning transformations in phase space that satisfy appropriate properties. Recall that if $\boldsymbol{\phi}(\bu)$ is a map in the phase space of equation \eqref{canonical_system} with $\bu = (\bq, \bp)$, then this  map is called symplectic if 
\begin{equation}
\left( \pp{\boldsymbol{\phi}}{\bu} \right)^T 
\mathbb{J}
\left( \pp{\boldsymbol{\phi}}{\bu} \right) = 
\mathbb{J} \, . 
\label{symplectic_map_def}
\end{equation}
A well-known result of Poincar\'e states that the flow $\boldsymbol{\phi}_t(\bu)$ of the canonical Hamiltonian system \eqref{canonical_system}, sending initial conditions $\bu$ to the solution at time $t$, is a symplectic map \citep{arnol2013mathematical,MaRa2013}. Several authors pursued the idea of searching directly for the symplectic mappings obtained from the data, instead of finding actual equations of the canonical systems and then solving them.

Perhaps the first work in the area of computing the symplectic transformations directly was done in  \citep{chen2020symplectic}, where \emph{Symplectic Recurring Neural Networks (SRNNs)} were designed. The SRNNs computed an approximation to the symplectic transformation from data using the appropriate formulas for symplectic numerical methods. An alternative method of computation of canonical Hamilton equations for non-separable Hamiltonians was done in \citep{xiong2020nonseparable}, building approximations to symplectic steps using a symplectic integrator suggested in \citep{tao2016explicit}. This technique was named \textit{Non-Separable Symplectic Neural Networks (NSSNNs)}.

A more direct computation of symplectic mappings was done using three different methods in \citep{jin2020sympnets,chen2021data}. 
The first approach derived in \citep{jin2020sympnets} computes the dynamics through the composition of symplectic maps of certain type, which was implemented under the name of \emph{SympNets}. Another approach \citep{chen2021data} derives the mapping directly using a generating function approach for canonical transformations, implemented as \textit{Genearating Function Neural Networks (GFNNs)}. The approach using GFNNs allows for an explicit estimate of the error in a long-term simulation. In contrast, error analysis in SympNets focuses on the local approximation error.   Finally, \cite{burby2020fast} developed \emph{H\'enonNets}, Neural Networks based on the H\'enon mappings, capable of accurately learning Poincar\'e maps of a Hamiltonian systems while preserving the symplectic structure.  SympNets,  GFNNs and H\'enonNets showed the ability to accurately simulate long-term behavior of simple integrable systems like a pendulum or a single planet orbit; and satisfactory long-term simulation for chaotic systems like the three-body plane problem.  Thus, learning symplectic transformations directly from data shows great promise for long-term simulations of Hamiltonian systems. 

The method of SympNets  was further extended for non-canonical Poisson systems by transforming the non-canonical form to local canonical coordinates using the Lie-Darboux theorem and subsequently using SympNets \citep{jin2022learning} by assuming that the dynamics occurs within a neighborhood in which the Poisson structure has constant rank. This method was named \emph{Poisson Neural Networks (PNNs). } While this method can in principle treat any Poisson system by learning the transformation to the canonical variables and its inverse, we shall note that there are several difficulties associated with this approach:
\rem{%%%BEGIN REM 
\todo{ \color{red}FGB: above is just to recall that they use the easy version of Darboux-Lie theorem which works in the neighborhood of points where the rank of the Poisson structure is constant. Around a point in which the rank is not constant (where symplectic leaves of various dimensions meet), one has more elaborate local forms involving a transverse Poisson structure (this was one of the confusion I had earlier regarding point 2 below). You can erase the box.}
} %%END REM 
\begin{enumerate} 
\item 
It is known that the Lie-Darboux transformation, in general, is only local: a global function transforming the system to canonical coordinates may not exist, although such transformation did exist for all examples presented in \citep{jin2022learning}. When such non-locality happens, on would need to define several transformations in overlapping domains and ensure the smoothness between them. It is not clear how the network-based Lie-Darboux function will perform in that case. 
\item  If an \emph{absolutely accurate} Lie-Darboux transformation coupled with a symplectic integrator was used to transform the coordinates to canonical form in \citep{jin2022learning}, it  would of course preserve all the Casimirs. However, any numerical errors in determining that transformation will yield corresponding errors in the Casimir evolution. 
%\todo{ \color{blue}FGB: Not sure what we have in mind here. We have a noncanoncal Poisson system, then we use the Lie-Darboux transformation to send it to some canonical form, then we apply a symplectic integrators in these local coordinates? The system is still not symplectic, no I am not sure what we do. Maybe we have in mind that we look at symplectic leaves (e.g. the coadjoint orbits), and apply the symplectic integrator directly there? (ignoring the other leaves)}
\item Since the errors in Casimir evolution are determined by the errors in the Lie-Darboux mapping, it is also not clear how these errors will accumulate over long-term evolution of the system. 
\end{enumerate} 
The preservation of Casimirs is especially important for predicting the probabilistic properties for the long-term evolution of many trajectories in the Poisson system \citep{Dubinkina2007}. In particular, even when the errors in each individual component of the solution may accumulate over time, the fact that the solution stays exactly on the Casimir surface will play an essential role in the probability distribution in phase space. 

%The theory behind \emph{PNN} limited the application to learning as single trajectory, although it was conjectured that the applications of the theory extend beyond these limitations. 

The  SympNets and PNN approach was further extended in \citep{bajars2023locally} where volume-preserving neural networks \emph{LocSympNets} and their symmetric extensions \emph{SymLocSympNets} were derived, based on the composition of mappings of certain type. A consistent good accuracy of long-term solutions obtained by  LocSympNets  and SymLocSympNets was demonstrated for several problems, including a discretized linear advection equation, rigid body dynamics and a particle in magnetic field. Although the methods of \citep{bajars2023locally} did not explicitly appeal to the Poisson structure of equations, the efficiency of the methods was demonstrated as applied to several problems that are essentially Poisson in nature, such as rigid body motion and the motion of particle in a magnetic field. However, the extension of the theory to more general problems was hampered by the fact that the completeness of activation matrices suggested in \citep{bajars2023locally} was not yet known. 

The limitations of the methods of \citep{jin2022learning} and \citep{bajars2023locally} come from the fact that they relate to a general system of equations, where it is assumed that very little is known about the general system apart the fact that it is Hamiltonian, or Poisson. On the other hand, there is a large class of physical systems where the Poisson bracket is known exactly; in particular, \emph{Lie-Poisson} systems. In these approaches, the bracket does not come as a consequence of the equations of motion, but from general considerations of the Lie group symmetries of the system. In that case, the Poisson bracket has an explicit expression stemming from the expression for the Lie algebra bracket, and is called the Lie-Poisson bracket.  The choice of actual Hamiltonian is a secondary step coming from physics. For example,  the motion of the rigid body comes from the invariance of the Lagrangian/Hamiltonian in the body frame with respect to rigid rotations, \emph{i.e.} ${\rm SO}(3)$ symmetry, see the introduction to the theory in Section~\ref{sec:intro_EP} below.   Table~\ref{Table_Lie_Poisson_examples} provides an incomplete list of physical examples admitting a description through Lie-Poisson or closely related brackets. 
\begin{table}[h!]
\centering
\begin{tabular}{|c|  c|}
\hline 
Problem &   Reference 
\\
\hline 
Rigid body &    \citep{holm2009geometric} \\
           & \citep{MaRa2013}
\\
\hline
Heavy top & \citep{holm2009geometric} \\ & \citep{MaRa2013}
\\
\hline
Underwater vehicles  & \citep{leonard1997stability}\\  & \citep{LeMa1997stability}\\ & \citep{holmes1998dynamics}
\\
\hline 
Plasmas &  \citep{morrison1980maxwell},\\ & \citep{marsden1982hamiltonian},\\ & \citep{holm1985nonlinear}, \\ & \citep{holm2010euler}
\\
\hline 
Fluids & \citep{MaWe1983},\\ &
\citep{MaRaWe1984},\\ &
\citep{holm1985nonlinear},\\ & \citep{morrison1998hamiltonian},\\ & \citep{morrison2006hamiltonian},\\
\hline 
Geophysical fluid dynamics & \citep{weinstein1983hamiltonian},\\\ & \citep{holm1986hamiltonian},\\ & \citep{salmon2004poisson}
\\
\hline 
Complex and nematic fluids & \citep{holm2002euler}, \\ & \citep{gay2009geometric},\\ & \citep{gay2010reduction}
\\
\hline 
Molecular strand dynamics & \citep{ellis2010symmetry},\\ & \citep{gay2012exact}
\\
\hline 
Fluid-structure interactions & \citep{gay2019geometric} 
\\
\hline 
Hybrid quantum-classical dynamics & \citep{GBTr2022},\\ &
\citep{GBTr2023}
\\
\hline 
\end{tabular}
\caption{A short description of some physical problems that can be written in Lie-Poisson form and related Poisson brackets.}
\label{Table_Lie_Poisson_examples}
\end{table}

In this paper, we develop the methods of constructing the activation maps directly from the brackets, predicting dynamics built out of maps computed by particular explicit solutions of Lie-Poisson systems. The method is applicable to \emph{all} finite-dimensional Lie-Poisson systems, as well as any Poisson system where explicit integration of appropriate equations for appropriate transformations are available.  The advantage of utilizing explicit integration in Lie-Poisson equations to drastically speed up calculations on every time step was already noticed in \citep{mclachlan1993explicit}, although the application was limited to Hamiltonians of certain form depending on the Poisson bracket. Our method is also applicable to arbitrary Hamiltonians; in fact, the Hamiltonian does not need to be known for the construction of the Neural Network. However, it is assumed that the underlying symmetry and the appropriate Lie-Poisson bracket are known. This is indeed often the case, as the detailed expression of the Hamiltonian in terms of the variables is often only approximate and is driven by the modeling choices for the particular physical system \citep{Tonti2013}, in contrast to the Lie-Poisson bracket itself. 

The novel contributions of the paper are as follows: 
\begin{enumerate}
\item 
We show that a large class of Poisson systems, namely Lie-Poisson systems obtained by symmetry reduction, allow explicit construction of maps for every particular Lie-Poisson bracket. 
\item These maps are global, in contrast to Lie-Darboux coordinates that are only local and may need to be re-computed depending on the position of solution in the configuration manifold. 
\item By construction, these maps preserve Casimirs of each particular bracket exactly, which is not necessarily the case for PNNs, LocSympNets, SymLocSympNets and any other methods known to us. 
\end{enumerate}

%The last point is important for data-based computations in continuum mechanics, since, as a rule, the spatial semi-discretization of a Poisson bracket for continuum systems does not yield a systems of ODEs that is Hamiltonian. Instead, a system is obtained that does not  satisfy Jacobi identity, but otherwise obtains all other properties of a Hamiltonian system (often termed quasi-Hamiltonian in the literature). This feature of discretization can be demonstrated explicitly in the example of two-dimensional shallow water \citep{salmon2004poisson,salmon2007general}. 

%\todo{Maybe add references to the few very special cases where Hamiltonian ODEs are obtained? And some discussion about how this is largely impossible in the general case?}

%The Jacobi identity is explicitly required for conservation of the bracket, symplecticity and phase volume \citep{MaRa2013}. As we will show, for some (but not all) continuum systems, our method will allow to evaluate the original Poisson bracket at discrete points, rather than facing the challenging task of discretizing the Poisson bracket, which may be even impossible. 

\section{An introduction to the Lie-Poisson equations} 
\label{sec:intro_EP} 
\subsection{General introduction: from Lagrangian to Hamiltonian description}\label{subsec_intro}

We start with a brief introduction of the origin of Lie-Poisson systems to introduce some notations and show why this particular type of approach is essential for many physical problems. We will consider only finite-dimensional systems here not to make the consideration too abstract.  Suppose a mechanical system is described by the coordinates $\bq$ and velocities $\dot \bq$, with $\bq$ lying on some configuration manifold $Q$ of dimension $n$. Hamilton's action principle is based on the Lagrangian function $L(\bq,\dot \bq)$ (possibly depending on time) and on the action $S=\int_{t_0}^{t_f} L(\bq, \dot \bq) \mbox{d} t$, and imposes the condition that the variations of the action must vanish on the variations of $\bq$ that are fixed on the boundaries $t=t_0,t_f$
\begin{equation}
\delta S = \delta \int_{t_0}^{t_f} L(\bq, \dot \bq) \mbox{d} t = 0 \, , \quad \de \bq (t_0) = \de \bq(t_f) =0 \, . 
\label{Hamilton_action_principle}
\end{equation}
In the Lagrangian approach, one takes the variations of \eqref{Hamilton_action_principle} and gets the Euler-Lagrange equations, which are second order equations in $\bq$. In the Hamiltonian approach, one introduces the momenta $\bp = \pp{L}{\dot  \bq}$ and assumes that this relation can be inverted for each $\bq$, giving the velocities as $\dot \bq = \dot \bq(\bq, \bp)$. One then defines the Hamiltonian function $H(\bp,\bq)=\bp \cdot \dot \bq(\bq,\bp) - L(\bq,\dot \bq(\bq,\bp))$, and the Euler-Lagrange equations of motion are equivalent to the \emph{canonical} Hamilton equations: 
\begin{equation}
\dot \bq = \pp{H}{\bp} \,  ,  \quad \dot \bp = - \pp{H}{\bq}\,. 
\label{canonical_eq}
\end{equation}
Any function $F(\bq,\bp)$ evolves according to the \emph{canonical bracket} 
\begin{equation} 
\frac{dF}{dt} = \left\{ F, H \right\} = \pp{F}{\bq}\cdot \pp{H}{\bp} - 
\pp{F}{\bp}\cdot \pp{H}{\bq}\,.
\label{canonical_bracket2}
\end{equation} 
One can see that the bracket \eqref{canonical_bracket2} satisfies the following properties that are valid for any functions $F(\bq,\bp)$, $G(\bq,\bp)$ and $H(\bq,\bp)$: 
\begin{enumerate}
\item Antisymmetry: $\{ F, H\} = - \{H , F\}$,
\item Linearity in each component: $\{ a F + b G, H \}=a \{ F, H \}+b \{ G, H \}$, $a,b \in \mathbb{R} $,
\item Leibniz rule (acts as a derivation): $\{ FG, H\} = F \{ G, H\} + G \{ F, H\}$,
\item Jacobi identity: $\{ F, \{ G, H\} \} + \{ H, \{ F, G\} \}+\{G, \{ H, F\} \} = 0$.
\end{enumerate}
After defining the total phase space $\bu=(\bq, \bp)$, the canonical bracket \eqref{canonical_bracket} and Hamilton's equations of motion can be written in coordinates as 
\begin{equation}
\{ F, H \} = \pp{F}{\bu}\cdot \mathbb{J}^{-1} \pp{H}{\bu} \, , \quad 
\dot{\bu} = \mathbb{J}^{-1} \pp{H}{\bu} \, , \quad 
\mathbb{J} = 
\left( 
\begin{array}{cc}
0 &  \mathbb{I}_n \\ 
-\mathbb{I}_n & 0 
\end{array}
\right) \, ,
\label{canonical_bracket_eq}
\end{equation}
where $\mathbb{I}_n$ is the $n \times n$ unit matrix. From the definition of $\mathbb{J}$ above, $\mathbb{J}^{-1}= \mathbb{J}^T=-\mathbb{J}$. 

\subsection{General Poisson brackets}
A general Poisson bracket $\{ F,H\}$ satisfies all four properties of the canonical Poisson bracket above, but cannot be necessarily expressed in coordinates as \eqref{canonical_bracket_eq}. Instead, the bracket and corresponding equations of motion are described more generally in local coordinates as  
\begin{equation}
\{ F, H \} = \pp{F}{\bu}\cdot \mathbb{B}(\bu) \pp{H}{\bu }\, , \qquad 
\dot{\bu} =  \mathbb{B}(\bu) \pp{H}{\bu}\, .
\label{gen_PB}
\end{equation}
In order for the bracket \eqref{gen_PB} to be Poisson, the matrix $\mathbb{B}(\bu)$, also known as the Poisson tensor, must be antisymmetric and satisfy a condition involving both the matrix $\mathbb{B}(\bu)$ and its derivatives. If such bracket can be found, the system is called (non-canonical) Poisson. 

A special attention should be paid to the case when the matrix $\mathbb{B}$ is degenerate. In that case, there very often\footnote{There are examples of degenerate Poisson brackets not admitting nontrivial Casimir functions, see \cite[Chap.10]{MaRa2013}.} are special functions for that particular bracket, called \emph{Casimirs}, which are conserved for \emph{any} Hamiltonian. By definition, a Casimir function $C$ satisfies
\begin{equation}
    \{ F, C\} = 0 \quad \mbox{for all }  F\, . 
    \label{Casimir_def}
\end{equation}
Any evolution must occur in such a way that all Casimirs of the system are conserved. Geometrically, the  motion is only possible on the intersection of Casimir level sets, no matter what the Hamiltonian for the system is.  

As we have seen, the motion of mechanical systems without friction is governed by canonical Poisson brackets, therefore the appearance of a non-canonical bracket \eqref{gen_PB} is somewhat mysterious. It turns out that in presence of symmetries, one can drastically reduce the degrees of freedom of the canonical Hamiltonian system by expressing the dynamics in terms of reduced coordinates, such as body or spatial coordinates. In such reduced variables, the dynamics is still governed by a Poisson bracket, but which is no more canonical. One of the most important example is the case of a system whose configuration is a Lie group: $Q=G$. Let us consider the case of a rigid body.
The configuration manifold is the Lie group of rotation matrices in $\mathbb{R}^3$, also known as $G=SO(3)$. This group consists of all $3 \times 3$ orthogonal matrices $\Lambda$ (\emph{i.e.}, $\Lambda^T \Lambda = \mathbb{I}_{3}$ with the determinant equal to $1$). The Lagrangian depends on the variables $(\Lambda, \dot \Lambda)$, and is just the kinetic energy. If one were to write either the Euler-Lagrange equations or the canonical Hamilton equations for $ \Lambda $ obtained by parameterizing $SO(3)$ using local representation of rotation matrices, such as rotation angles, one would end up with complicated and unwieldy equations. Instead, the equations of motion for the rigid body can be efficiently written in terms of the angular velocity in the body frame $\bOm$ using the tensor of inertia $\mathbb{I}$ as: 
\begin{equation} 
\mathbb{I} \dot \bOm = - \bOm \times \mathbb{I} \bOm  \, . 
\label{Rigid_body_eqs}
\end{equation} 
As it turns out, the equations \eqref{Rigid_body_eqs} can be understood from the point of view of symmetry. The rigid body kinetic energy is invariant with respect to \emph{left} rotations $\Lambda \rightarrow R \Lambda$, where $R \in SO(3)$ is a fixed rotation matrix. We can thus express the kinetic energy in terms of the antisymmetric $3\times 3$ matrices $\widehat{\Omega}= \Lambda^T \dot \Lambda$ which take on the role of angular velocities.  One can compute the vector $\bOm$ as $\widehat{\Omega}_{ij} = \epsilon_{ijk} \boldsymbol{\Omega } _k$, with $\epsilon_{ijk}$ being  the   Levi-Civita symbol. The equations of motion \eqref{Rigid_body_eqs} can be written in terms of momenta $\bPi$, with the utilization of the rigid body bracket: 
\begin{equation}
\dot \bPi = - \pp{H}{\bPi} \times \bPi \, , \quad \left\{ F, H\right\}:= - 
\bPi \cdot \left( \pp{F}{\bPi} \times \pp{H}{\bPi} \right) \, . 
\label{RB_equations}
\end{equation}
The bracket in \eqref{RB_equations}, as it turns out, satisfies the four properties of a Poisson bracket and is expressed in terms of the Lie bracket on the Lie algebra of $SO(3)$, given by the vector product, stemming from the invariance of the system with respect to $SO(3)$ rotations.

These ideas can be generalized as follows, see  \citep[Chap.10]{MaRa2013}. For a general system defined on a Lie group $G$, one naturally has the associated Lie algebra $\mathfrak{g}$ with Lie bracket denoted as $[\alpha, \beta]$ for $  \alpha , \beta \in \mathfrak{g} $. If $\left\{e_a\right\}$, $a=1, \ldots, n$ is a basis of $ \mathfrak{g} $, then the Lie bracket is locally expressed in terms of the structure constants $C^{d}_{ab}$ such that
\begin{equation} 
[e_a, e_b] = C^{d}_{a b} e_d. 
\label{Structure_const_def}
\end{equation}
Let us denote by $\langle \mu  , \alpha  \rangle$ the duality pairing between vectors $ \alpha $ in the Lie algebra  $ \mathfrak{g} $ and co-vectors (or momenta) $ \mu $ in the dual space $ \mathfrak{g} ^* $ to $ \mathfrak{g}$. The partial derivatives of functions $F, H: \mathfrak{g} ^* \rightarrow \mathbb{R}$ with respect to $\mu$ thus belong to $ \mathfrak{g}$, and one can define the \emph{Lie-Poisson bracket} derived from the Lie bracket as follows: 
\begin{equation}
\{ F, H \} = \pm \left< \mu , \left[ \pp{F}{\mu} \,, \pp{H}{\mu} \right] \right> \,.
\label{LP_bracket}
\end{equation}
We refer to \ref{app:appendix_geometric_LPNets} for the explanation of the $\pm$ sign and for the relation between this bracket and the canonical bracket.
In terms of coordinates, the bracket \eqref{LP_bracket} is a particular case of \eqref{gen_PB} with the matrix $\mathbb{B}({\mu})$ defined as 
\begin{equation} 
\mathbb{B}_{ab} (\mu) =  \pm  C_{ab}^d \mu_d 
\label{B_matrix_LP}
\end{equation} 
and the Lie-Poisson equations are expressed in coordinates as 
\begin{equation}
\dot \mu_a = \pm C^{d}_{a b} \mu_d \pp{H}{\mu_b}\,.
\label{eqs_Lie_Poisson}
\end{equation} 
One can verify that the Lie-Poisson bracket \eqref{LP_bracket} satisfies all the conditions of a general Poisson bracket. While the Lie-Poisson brackets appear because of the fundamental considerations of the symmetries of the physical system, the Hamiltonian is a modelling choice related to physics. Thus, we assume that if a system possesses a Lie-Poisson bracket, it is known explicitly \emph{a priori} and does not need to be computed or determined from the data. However, the dependence of the Hamiltonian on its arguments and parameters is not known and must be determined.

\section{LPNets as Poisson maps for a general system} 
\label{sec:general_LPNets}

We are now ready to consider the theory of  structure-preserving neural networks. We follow the ideas originally introduced in \citep{jin2020sympnets} as SympNets. The idea behind SympNets is to learn in the space of available  symplectic transformations for a canonical system. Further,  \citep{jin2022learning} introduced  a generalization of this method to an arbitrary Poisson system by using the Lie-Darboux theorem, stating that every Poisson system can be locally transformed into a canonical form,  implementing these ideas a PoissonNets. This theory was further extended in  \citep{patel2022thermodynamically,zhang2022gfinns} for thermodynamics systems using the GENERIC metriplectic bracket approach. 

Mathematically speaking, previous works in this avenue of thinking \citep{jin2020sympnets,jin2022learning} and \citep{bajars2023locally} seek to determine a mapping $\phi_h(\mathbf{y}_i$) in the next computational point $\mathbf{y}_{i+1}$, satisfying as many restrictions preserving the structure of the actual flow as possible. The authors developed universal mappings that are symplectic in the canonical coordinates $(\bq,\bp)$, and sought the solution as a combination of these mappings. Since the mappings are universal for all finite-dimensional Hamiltonian systems, one then needed to prove the convergence and completeness of the combinations of these mapping, especially when coupled to the Lie-Darboux theorem which maps the Poisson system to a canonical form. 

We build on the ideas of SympNets and Poisson nets by directly constructing  the transformations related to the Poisson systems (\emph{Poisson maps}) for the \emph{known} Lie-Poisson bracket, and using them as elements for construction of LPNets. 

We first recall the useful concept of \emph{Poisson maps}.

\begin{definition}\label{def_Poisson}\citep[\S 10.3]{MaRa2013} Let $(P_i, \{ \cdot , \cdot \}_i)$, $i=1,2$ be two Poisson manifolds. A mapping $f: P_1 \rightarrow P_2$ is Poisson if it preserves the Poisson bracket, \emph{i.e.}, for all functions $F,G:P_2 \rightarrow \mathbb{R} $
\begin{equation} 
\{ F,G \}_2 \circ f = \{ F \circ f, G \circ f \}_1\, . 
\label{Poisson_map_def} 
\end{equation} 
\end{definition}

A critical piece of information for our further progress is contained in the following result.

\begin{theorem}[Hamiltonian flows are Poisson]
\citep[Thm. 10.3.1.]{MaRa2013} Consider a Poisson manifold $(P, \{ \cdot , \cdot \})$ and a Hamiltonian $H:P \rightarrow \mathbb{R} $. Let $\phi_t(\bu_0)$ be the flow of the Poisson system associated with $H$, see  \eqref{gen_PB}, which maps the initial conditions $\bu_0$ to the solution $\bu(t)$ at time $t$. Then, the mapping $\phi_t$ is Poisson, \emph{i.e.}, it satisfies:
\begin{equation} 
\left\{ F \circ \phi_t, G \circ \phi_t \right\}= 
\left\{ F , G  \right\}  \circ \phi_t\,,
\label{Poisson_bracket_conservation}
\end{equation} 
for all functions $F,G: P \rightarrow \mathbb{R}$.
\end{theorem}

Following this theorem, we can design  Poisson maps for a particular Poisson bracket using a sequence of flows created by simplified Hamiltonians for the  Lie-Poisson dynamics. 
The advantage of this approach is that the resulting transformations will be Poisson and will be able to approximate any flow locally for the particular system considered. The disadvantage of our approach is inherently intertwined with the advantages: the mappings have to be constructed explicitly for every Lie-Poisson bracket. Fortunately, as we show below, this is possible as these mappings are derived as solutions of a linear system of ODEs. The method presented here is reminiscent of the Hamiltonian splitting methods used in numerical analysis \citep{mclachlan1993explicit,mclachlan2002splitting}, reformulated for the purpose of data-based computations and the use of neural networks to find appropriate parameters of the Hamiltonians.

\section{A general application to finite-dimensional Lie groups}
\label{sec:Lie_Group_LPNets}

The key to this paper lies in considering the evolution of the momentum $\mu$ coming from the equations \eqref{eqs_Lie_Poisson} for particular expressions for the Hamiltonian, namely, Hamiltonians linear in momenta 
\begin{equation} 
H( \mu )=\left< \alpha, \mu \right> = \alpha^a \mu_a\, , 
\label{LP_Hamiltonians}
\end{equation} 
where $\alpha^a$ are some constants that are to be found based on the learning procedure. 
The flow generated by the Hamiltonian \eqref{LP_Hamiltonians} is given by a linear equation in $\mu$ that can be written in two equivalent ways (choosing the + sign in \eqref{eqs_Lie_Poisson}): 
\begin{equation}
\dot \mu_a = C^d_{a b} \alpha^b \mu_d := \mathbb{M}(\alpha)^d_a  \mu_d = \mathbb{N}(\mu)_{a b}  \alpha^b
\, . 
\label{eq_LPNets}
\end{equation}
The number of possible dimensions of $\alpha$ is exactly equal to the dimension of the momentum space. However, the "effective" dimension of this space  may be less, and is related to the dimension of the Lie algebra $n$ minus the dimension of the kernel of  the operator $\mathbb{N}( \mu ): \mathfrak{g} \rightarrow \mathfrak{g} ^* $, see Remark \ref{remark_effective_dim}.  The operators $\mathbb{M}(\alpha)$ acting on the space of momenta $\mu$ and $\mathbb{N}(\mu)$ acting on the space of $\alpha$  can be described in the coordinate-free form as $\operatorname{ad}^*_\alpha \square $ and $\operatorname{ad}^*_\square \mu $, respectively, see  \ref{app:appendix_geometric_LPNets}.

Let us assume that there are $d$ independent Casimir functions $C_j( \mu )$, $j=1,...,d$. From \eqref{Casimir_def} and \eqref{B_matrix_LP} such functions satisfy $ C^d_{ab} \frac{\partial C_j}{\partial \mu _b} \mu _d=0$ for all $ \mu $, i.e., $\partial C_j/ \partial \mu $ must belong to the kernel of the operator $\mathbb{N}(\mu)$. We shall consider the effective dimension of the space of all possible $\alpha$ to be exactly $n-d$, where $n$ is the dimension of the momentum space and $d$ is the number of  independent Casimirs. Thus, locally the vectors $ \alpha $ in this effective space form exactly the right number of local tangent vectors to the intersection of Casimir surfaces to reach any point locally using a flow generated by the Poisson map. The next step is to compute that map exactly.

\begin{remark}[On the effective number of dimensions]\label{remark_effective_dim}
In general, the number of null directions of $\mathbb{N}(\mu_0)$, denoted as $k(\mu_0)$ may depend on the momenta $ \mu_0 \in \mathfrak{g} ^*  $ (or, more precisely, on the actual coadjoint orbit the solution is on), whereas the number of independent Casimirs $d$ is fixed. The dimension of the image of the map $ \xi \in \mathfrak{g} \mapsto \mathbb{N}( \mu _0) \xi =\operatorname{ad}_ \xi^* \mu _0 \in \mathfrak{g} ^* $ is the dimension of the coadjoint orbit of $ \mu _0$, denoted as $\mathcal{O} _{ \mu _0}$. Thus, in general, we have $ k( \mu _0)+ \operatorname{dim}( \mathcal{O} _{ \mu _0})= n$ and we have $d\leq k( \mu _0)$ for almost all $ \mu_0$, but not necessarily $d= k( \mu _0)$\;\footnote{Note that there are examples of Lie algebras whose generic coadjoint orbits have codimension strictly bigger than the number of independent Casimirs.}. However, the data are exceptionally unlikely to lie on orbits with high codimension $k(\mu) > d$, and we are going to assume that the effective number of dimensions of $\alpha$ is $n-d$. If the data was obtained from one of the orbits with high codimension, one would amend formula \eqref{alpha_mod_Casimirs} below using the information about that exceptional orbit. Note that our method preserves the general form of coadjoint orbits in all cases, whether the orbit is exceptional or not. This fact could be useful in data-based computations of exceptional orbits. It is an interesting question  which we will address in the follow-up work.
\end{remark}

Equations \eqref{eq_LPNets} are linear differential equations in $\mu$, and the solutions of these equations can be (in principle) found in explicit form. 
This solution can be written in compact form as 
\begin{equation} 
\mathbb{T}(t,\alpha) \mu_0 = e^{\mathbb{M}(\alpha) t} \mu_0 \, . 
\label{LPNets_transformation}
\end{equation} 
%\todo{FGB: Technically, the solution is
%\[
%\operatorname{Ad}^*_ {\mathbb{T}(t,\alpha) ^{-1} } \mu _0
%\]
%so the notation in \eqref{LPNets_transformation} should be understood as a coadjoint action. I agree that we don't want to use $ \operatorname{Ad}^*$ here, but is it clear to the reader what is meant in \eqref{LPNets_transformation}? (knowing that even for matrices this won't take the form $e^{\mathbb{M}(\alpha) t} \mu_0 $ but rather something like $e^{\mathbb{M}(\alpha) t} \mu_0 e^{-\mathbb{M}(\alpha) t}$ (only for $SO(3)$ it actually looks like \eqref{LPNets_transformation}).
%}
The mappings $\mathbb{T}(t,\alpha)$ defined in \eqref{LPNets_transformation} satisfy all the requirements as the building blocks for the neural networks.

Let us denote the mapping $\mathbb{T}_a(t, \alpha^a)$, $a=1,...,n$, to be the map originating from the $a$-th component of $\alpha$ to have the value of $\alpha^a$ and all other values being zero. We divide each time step into $n$ substeps $h_a$ with $\sum_a h_a =h$. Notice that  the number of steps is $n$ and not $n-d$, where $d$ is the number of Casimirs, as we explain later.

These mappings satisfy the following conditions: 
\begin{enumerate}
\item The mappings $\mathbb{T}_a(h_a,\alpha^a)$ are Poisson, see \eqref{Poisson_bracket_conservation}, for the Lie-Poisson bracket,
\item The mappings conserve all Casimirs of the system, \footnote{ And, in fact, these maps also preserve all types of coadjoint orbits, \emph{i.e.}, they accurately represent the coadjoint action ${\rm Ad}^*_g \mu$. This information would be useful if we were to compute exceptional orbits, which we will not do here.} 
\item Any two points close enough to each other on the same Casimir surface can be connected using a combination of mappings $\mathbb{T}_a(h_a,\alpha^a)$.
\end{enumerate}

The procedure of constructing LPNets is as follows: 
\begin{enumerate} 
\item Find explicit solutions for \eqref{LPNets_transformation} by solving equations \eqref{eq_LPNets}. 
\item Define $\bar \alpha = (\alpha^1, \ldots, \alpha^n)$ and
\begin{equation} 
\mathbb{T}(\bar \alpha) = \mathbb{T}_n(h_n, \alpha^n) \circ \mathbb{T}_{n-1}(h_{n-1}, \alpha^{n-1}) \circ \ldots \circ \mathbb{T}_1(h_1, \alpha^1)\,.
\end{equation}
%\todo{ \color{red} FGB: This is a tiny detail, but these factors $\mathbb{T}_a(h_a, \alpha^a)$ don't commute, so $\mathbb{T}(\bar \alpha)$ may slightly change if we change the order of coordinates. I guess this doesn't change anything for the results, right?.\\
%Any reason why not just choose all the substeps $h_a$, to be equal to $h/n$? \\ 
%\textcolor{magenta}{Correct, if one changes the order of operators then $\alpha$ would change as well. Regarding the equipartition of steps, I actually tried to have another parameter for one of the problems (don't remember which one) in order to get higher accuracy and didn't get very far. But it doesn't mean that it is a useless endeavor - I put a short sentence here.  }}
For the set of $N$ data pairs $(\mu_i^0, \mu_i^f)$, $i=1, \ldots N$ set up a minimization procedure to find the set of numbers $\bar \alpha$ for each $\mu_i^0$, minimizing the ``individual" square loss 
\begin{equation}
\bar{\alpha}_i= \mbox{arg min} \left| \mathbb{T}(\bar{ \alpha_i}) \mu_{i}^0  - \mu_{i}^f \right|^2
\label{Loss_function}
\end{equation}
In most of the examples we consider here, we can find $\bar \alpha$ analytically; however, in more general situations, one could also utilize a root-finding or a minimization procedure finding $\bar \alpha_i$ for every pair of data points. 
\item Since $\bar \alpha$ are only defined up to a vector normal to the Casimir surface, in order to make the data consistent, we need to project out the appropriate components of the Casimirs. We will need to find coefficients $p_j$, $j=1, \ldots, d$ such that  the projection of $\bar \alpha$ on the gradients of the Casimirs vanishes: 
\begin{equation} 
\bar \alpha_i \rightarrow  \bar \alpha_i - \sum_{j=1}^d p_j \left. \pp{ C_j}{\mu} \right|_{\mu=\mu_i^0} \, ,  \mbox{  with }
\left< \bar{\alpha}_i \,, \, \left. \pp{ C_j}{\mu} \right|_{\mu=\mu_i^0}\right>  =0  \, ,j=1,...,d,
\label{alpha_mod_Casimirs}
\end{equation} 
 where $C_j$, $j=1,...,d$ are a set of $d$ independent Casimir functions.
\item Create a neural network approximating the mapping $\mu_i \rightarrow \bar \alpha_i$. This function will be denoted as $\bar \alpha= NN(\mu)$. 
\end{enumerate} 
Since the operators $\mathbb{T}_a$, in general, do not commute, the composition order should be fixed ahead of time and also be the same for all data pairs. A different choice of the composition order will lead to a different (but of course equivalent) set of parameters $\alpha$. The choice of the composition order of $\mathbb{T}_a$ has to be preserved in the prediction step as well. Also, we do not consider different time sub-steps $h_a$ here, taking them all to be equal.  It is possible that the choice of $h_a$ can be made to improve accuracy or convergence properties of the scheme. 

The solution starting at a given initial condition $\mu_0$ can then be evaluated using the neural network. If $\mu=\mu_j$ at the time step $j$, then the value of the solution at the next step is computed as 
\begin{equation} 
\mu_{j+1} = \mathbb{T}(\bar \alpha_j^{\rm est}) \mu_j , \quad \bar\alpha_j^{\rm est} = NN(\mu_j)\, . 
\label{sol_eval_iteration}
\end{equation} 
Note that we never compute the actual Hamiltonian, its gradients, or the equations of motion. Instead, we just compute the composition of Poisson transformations reproducing the dynamics in phase space of some unknown Poisson system -- with the known Lie-Poisson bracket. 

We will now apply this procedure to several particular examples of physical systems which have a Lie-Poisson bracket. We shall call this method Local \emph{LPNets}, or simply LPNets, as the exponential representation of the map is only defined locally.  The advantage of this method is that when operating on Lie groups, we are guaranteed to be able to reach any point locally with an exponential map, so the completeness is automatic. However, this mathematical simplicity has to be compensated by the necessity to apply a neural network to learn the mappings from data points in the neighborhood of a trajectory. After our description of LPNets, we develop a more complex procedure which derives Lie-Poisson activation modules that directly extend the work of \citep{jin2020sympnets,jin2022learning,bajars2023locally}. We call these methods global Lie-Poisson networks, or G-LPNets. Using the example of a rigid body motion, we show that G-LPNets provide a promising simple, accurate and highly computationally effective neural network.  

A more general derivation performed in the coordinate-free form and in the general language of modern geometric approach is presented in \ref{app:appendix_geometric_LPNets}. 

%\todo{VP: These came out of Francois' question about nonlinear mapping (see below). I couldn't prove completeness but in general the numerical results are outstanding. If one could prove completeness of these mappings, even for several particular Lie groups, it would be a fantastic result. Maybe Sofiia can solve it in her PhD? }

%\begin{framed} 

%\todo{\textcolor{red}{ FGB: Maybe there are Lie group analogue to the linear/activation/gradient modules of  \citep{jin2020sympnets}. They correspond to Hamiltonians of the form $H(q,p)= V(q)$ and $H(q,p)= V(p)$. Unfortunately these forms can never be $G$-invariant on $T^*G$.\\
%In this case a nonlinear Hamiltonian is needed at some point to prove the approximation properties (the activation or gradient modules). So, I am not sure if using only linear Hamiltonians is enough. Maybe this is a hard question?
%Finally, maybe we can obtain something from the dimensions of the coadjoint orbits.}
%\\ 
%\textcolor{magenta}{Yes, indeed! This is now presented as G-LPNets, and it works astonishingly good, but we still need to think about the completeness... }}
%\end{framed} 

\section{Test cases for LPNets}

We choose to use the same test cases as in \citep{jin2022learning,bajars2023locally}, except for an additional test for $SE(3)$ (the underwater vehicle). No detailed comparisons are performed between our results and those in \citep{jin2022learning,bajars2023locally}. This is because the accuracy of each method depends strongly on the structure of the neural network, the accuracy and distribution of data used in training, and specific learning procedures employed, which prevents a fair comparison between methods. Instead, we will outline the performance of LPNets on these problems, and contrast with general results from \citep{jin2022learning,bajars2023locally}. In all cases, LPNets conserve the Casimir functions to machine-precision, unlike the other approaches.

In what follows, we will develop the prediction of parameters for LPNets using a dense Neural Network structure. Naturally, such a dense network will require appropriate number of data points to avoid overfitting. In Section~\ref{sec:G_LPNets}, we show how to reduce the size of that network and achieve the accuracy in the whole phase space with a particular structure of a network which we will call G-LPNets.
\subsection{Rigid body dynamics} 
\label{sec:rigid_body}
\paragraph{Motion of a rigid body  as dynamics on the Lie group $SO(3)$} The (Lie-)Poisson bracket, the Hamiltonian, and the corresponding equations of motion for momenta $\bPi$ (measured from the body frame of reference), are \citep{holm2009geometric}: 

\begin{equation} 
\begin{aligned} 
\left\{ F, G \right\} & = - \bPi \cdot \left( \pp{F}{\bPi} \times \pp{G}{\bPi} \right) 
\\
H(\bPi)& =\frac{1}{2} \bPi \cdot \mathbb{I}^{-1} \bPi 
\\
\dot \bPi & = - \mathbb{I}^{-1} \bPi \times \bPi \,.
\end{aligned} 
\label{Rigid_body_LP} 
\end{equation} 
Suppose there is a sequence of pairs of initial and final points of transformation $(\bPi_i^0,\bPi_i^f)$, $i=1, \ldots, N$, coming from some information that we call \emph{ground truth}.  If that sequence comes from a single trajectory of length $N$, \emph{i.e.} has the form $\bPi_0, \bPi_1, \ldots, \bPi_N$ we take  $\bPi_i^0=\bPi_i$, $\bPi_0^f = \bPi_{i+1}$. However, our method does not explicitly assume the existence of a single trajectory for learning. 

To find the Poisson map approximating  the motion of the system at every time step $i$, let us consider Hamiltonians $H_i$ linear in momenta, \emph{i.e.} having the form $H_i(\bPi)=\mathbf{A}_i \cdot \bPi$, where $\mathbf{A}_i$ is some unknown constant vector that is different for every pair of points.  The Hamiltonian flow generated by that Hamiltonian is described by 
\begin{equation} 
\dot \bPi = - \pp{H_i}{\bPi} \times \bPi = -  \mathbf{A}_i \times \bPi \, . 
\label{Euler_Simple_Hamiltonian} 
\end{equation} 

Note that we can also consider a Hamiltonian of the form $H(\bPi)=f(\mathbf{A} \cdot \bPi)$. The dynamics induced by this Hamiltonian preserve the quantity $\mathbf{A}\cdot \bPi$, as one can see from \eqref{Euler_Simple_Hamiltonian}. Thus, that extension corresponds to a simple redefinition of $\mathbf{A}$ by scaling and does not bring extra insight into the problem.\footnote{Of course, $\mathbf{A} \cdot \bPi$ is not a constant for the general system \eqref{Rigid_body_LP} - that is only true for the particular choice of the Hamiltonian $H(\bPi)=f(\mathbf{A}\cdot \bPi)$.} Thus, the motion defined by \eqref{Euler_Simple_Hamiltonian} is a rotation of a vector $\bPi$ about the axis $\mathbf{A}_i$ with a constant angular velocity. The flow preserves the Lie-Poisson bracket since it is a Hamiltonian flow with the same bracket. The dynamics \eqref{Rigid_body_LP} also preserve the Casimir $C(\bPi)=|\bPi| ^2 $ exactly. After the time $t=h$, the dynamics \eqref{Rigid_body_LP} rotates the vector of angular momentum $\bPi$ by an angle $\phi(\mathbf{A}_i)=|\mathbf{A}_i| h$ around the axis $\mathbf{n}_{\mathbf{A_i}}=\mathbf{A}_i/|\mathbf{A}_i|$. In other words, if $\mathbb{R}(\mathbf{n},\phi)$ is the matrix of rotation around the axis $\mathbf{n}$ by the angle $\phi$, then \eqref{Rigid_body_LP} transforms the momentum as $\bPi \rightarrow \mathbb{R}(\mathbf{n}_{\mathbf{A_i}},\phi(\mathbf{A}_i)) \bPi$. 

Everywhere in this paper, the ground truth is obtained by BDF integrator in Python's \emph{Scipy} package, with relative and absolute tolerances being set at $10^{-13}$ and $10^{-14}$, respectively. Note that this numerical method is not expected to conserve Casimirs so our method will actually be more precise than the ground truth. We note that while Lie-Poisson integrators preserving Lie-Poisson structure do exist \citep{marsden1999discrete}, the precise implementation needs to be rederived in an explicit form for each particular Lie-Poisson system. We found it to be more appropriate to use a high accuracy algorithm that is common to all problems for a fair comparison, rather than build an algorithm that is tailored to every particular problem. The accuracy of the BDF algorithm is more than sufficient for our ground truth calculation and providing comparisons with the previous works.  
%\todo{\color{blue} FGB: Actually there are such variational integrators preserving the Lie-Poisson structure, see \url{https://arxiv.org/abs/math/9909099}, (but not preserving the Casimirs), so better not saying this.} \\ 
%\textcolor{magenta}{VP: Ah.. Interesting, thanks for the reference. I think it would be a good paper to compare to. They seem to claim they preserve the Casimirs, at least in the Nonlinearity paper. I have changed the text above to show that we are aware of their paper, and changed to the fact that we want to compare to something common that is easy to use, rather than separate thing that is redone for every particular problem.  They also talk about the Hamiltonian splitting methods which is pretty close to what we are discussing (without any neural networks of course). }
Thus, to be consistent, we used a high precision non-symplectic integrator for all cases, carefully checking its accuracy in all applications.

\paragraph{Data preparation} 
We should consider three angles of rotations. Given a sequence of begin and end pairs of momenta $(\bPi_i^0,\bPi_i^f)$, we compute $
\mathbf{A}_i$ as a function of $\bPi_i^0$ as follows: 
\begin{equation} 
\mathbf{A}_i = \frac{1}{h} \bPi_i^0 \times \bPi_{i}^f
\label{SO3_A_def}
\end{equation} 
and the angle $\theta_i$ of rotations from vector $\bPi_0^i$ and $\bPi_f^i$ as the shortest motion along the sphere $\bPi=$const. The cross product contains information for both the direction normal to both $\bPi_i^0$ and $\bPi_{i}^f$, and the angle of rotation, as described above. The factor $1/h$ is introduced to normalize the output data to be of order 1. 

\begin{figure} 
\centering 
\includegraphics[height=0.26 \textheight]{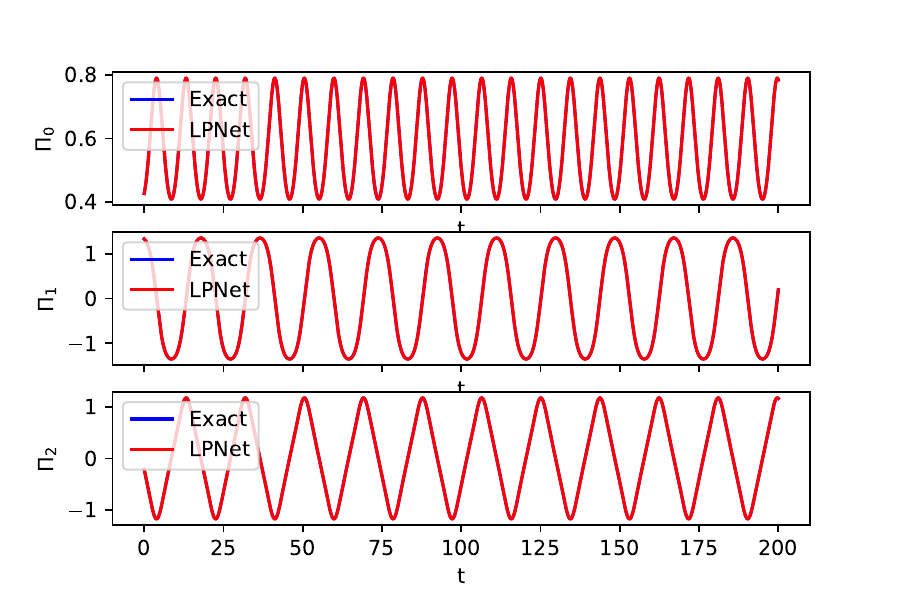}
\includegraphics[height=0.26 \textheight]{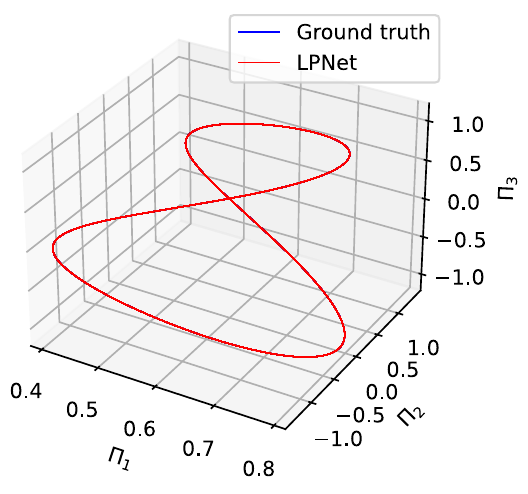}
\caption{Left: Results of LPNets applied to the motion of a rigid body (red) versus ground truth (blue) for the individual momenta. Right: Parametric plot of the momenta in the phase space. The results are visually indistinguishable.  \label{fig:LP_results_rigid_body} }
\end{figure}

We could of course get $\mathbf{A}_i$ by finding the match of three subsequent rotations around, say, Euler angles, and subtracting the corresponding rotation about the axis $\bPi_i^0$ or $\bPi_i^f$,  applying equation  \eqref{Loss_function} directly. 
However, the notation of vector cross product, only available for $SO(3)$, provides a simple and efficient alternative to this more complex procedure. 
\paragraph{Neural network}
Make a standard neural network learning from the data for the mapping $\bPi \rightarrow \mathbf{A}$. The neural network will have the three components of $\bPi$ as inputs and the three components of $\mathbf{A}$ as outputs. Here and everywhere else, we utilize the package \emph{Tensorflow}\footnote{\url{https://www.tensorflow.org}}. 
In the results shown in Figure~\ref{fig:LP_results_rigid_body}, the learning is done on $N=1000$ pairs produced by the single trajectory originating at $\bPi_0 = (1/\sqrt{2},-1/\sqrt{2},1)$, with the interval between trajectory points given by $h=0.1$. 

The neural network has three hidden layers of $16$ neurons with the sigmoid activation function, with $659$ trainable parameters. Out of $1000$ pairs as the input data, $80\%$ are used for training and $20\%$ for evaluation, with the loss measured as the mean square discrepancy between $\mathbb{R}(\mathbf{n}_{\mathbf{A_i}},\phi(\mathbf{A}_i)) \bPi_i^0$ and $\bPi_i^f$. Adam optimization algorithm is used with the learning rate starting at $10^{-3}$. The loss and validation loss reach the values of approximately $3.8 \cdot 10^{-9}$ and $4.1 \cdot 10^{-9}$ after $10^{5}$ epochs. 
 
\begin{figure} 
\centering 
\includegraphics[width=0.48 \textwidth]{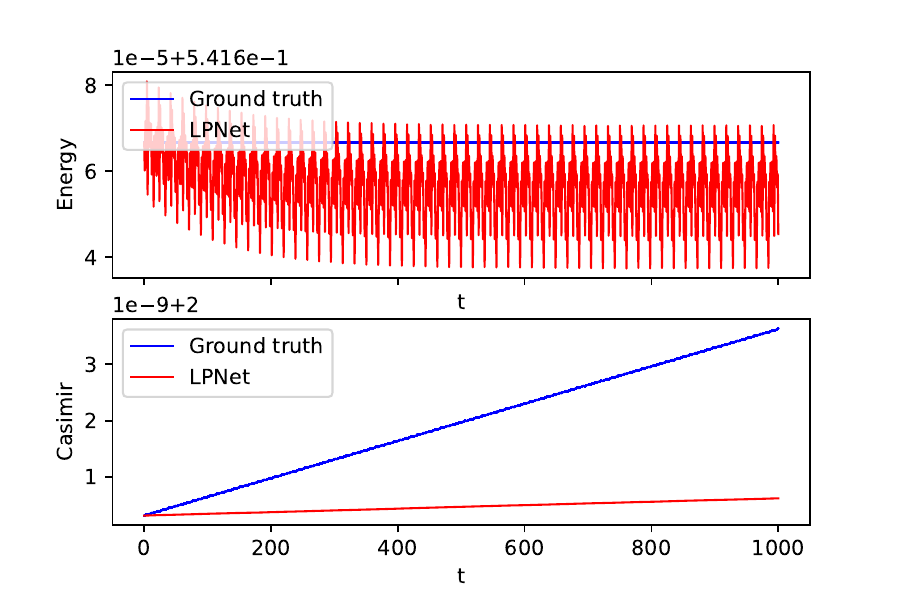}
\includegraphics[width=0.48 \textwidth]{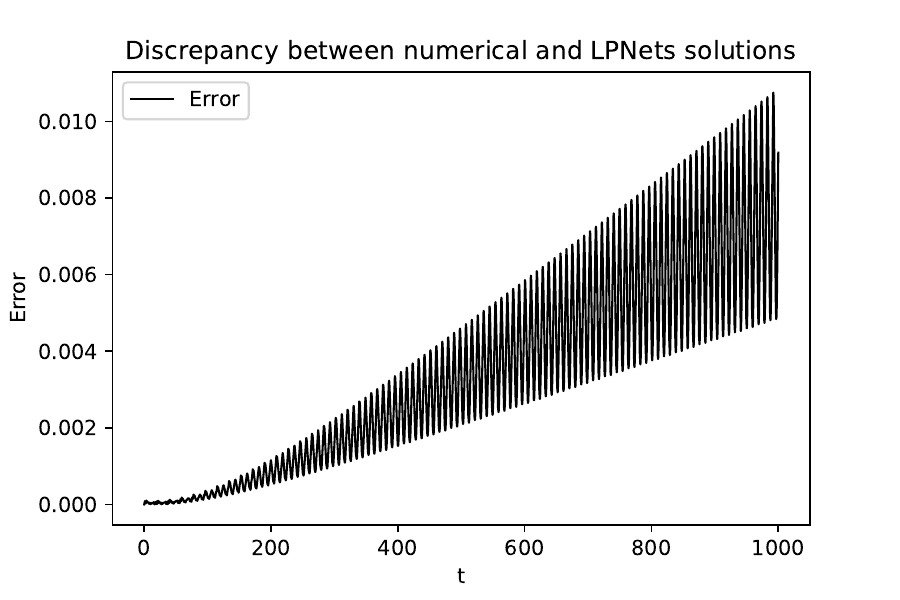}
\caption{Left: Conservation of the Hamiltonian $H$ (top) and the Casimir $C$ (bottom), comparing the results of LPNets (red) and ground truth (blue). Notice that LPNets conserve the Casimir exactly (to machine precision) and thus substantially exceeds the ground truth in the conservation of Casimirs. Right: The discrepancy between the results of LPNets and the ground truth. The discrepancy comes mostly from time mismatch, whereas the amplitude of oscillations is conserved with high precision.   \label{fig:LP_Conservation_error_rigid_body} }
\end{figure}

\paragraph{Prediction}  
The predicted trajectory start $\bPi_0$ is taken to coincide with the endpoint of the learning trajectory, with the values of $ \bPi_0 \simeq (0.43, 1.33,-0.21)$.  On each step, once $\bPi_{j-1}$ is known, Neural Network creates the prediction for the rotation axis $\mathbf{A}_j/h$ and the angle $\phi_j = |\mathbf{A}_j|/h$. That prediction of the neural network is used to produce the mappings $\bPi_{j}= \mathbb{R}(\mathbf{A}_j,\phi_j) \bPi_{j-1}$, where  $j=1 \ldots m$ after the desired number of steps. That prediction by LPNets is compared with the ground truth prediction obtained by high accuracy ODE solver as described above. We perform $10000$ time steps to reach $t=1000$ and present the results in Figure~\ref{fig:LP_results_rigid_body} and Figure~\ref{fig:LP_Conservation_error_rigid_body}. Note that for clarity, only the first $2000$ time steps up to $t=200$ are shown in the individual momenta plots, the left panel of Figure~\ref{fig:LP_results_rigid_body}. The right panel of that Figure shows that all available data coincide perfectly. 

In Figure~\ref{fig:LP_Conservation_error_rigid_body}, we present the results for the conservation of the Hamiltonian $H(\bPi)$ and the Casimir $C(\bPi)=| \bPi|^2$. The Hamiltonian is preserved to about $0.01\%$ relative accuracy. The Casimir in the ground truth solution is preserved to about $10^{-9}$ accuracy. In the LPNets solution, the Casimir is preserved to machine precision, far exceeding possible accuracy for ground truth for this variable. 
On the right side of this panel, we plot the error of the solution as the $L_2$ norm of deviation between two solutions (ground truth and LPNets) for each $t$. The deviation is growing roughly linearly in time to the values of about $0.5$ after the time $t=1000$. This may come as a surprise given the excellent agreement on the right hand side of the Figure~\ref{fig:LP_results_rigid_body} for all $t$. This phenomenon has a simple explanation: high accuracy of conservation laws presented on Figure~\ref{fig:LP_Conservation_error_rigid_body} forces the solution to exist on the intersection of $H=$const (an ellipsoid) and $C=$const (a sphere) with the high accuracy.

%The discrepancy then comes from the effective value of the time step that is slightly different in LPNets and the ground truth. The right side of the panel on Figure~\ref{fig:LP_results_rigid_body} presents essentially that intersection of two surfaces, and is thus satisfied to a very high precision. 

\paragraph{Comparison with the previous literature} The case of rigid body motion was considered in \citep{bajars2023locally}. The appropriate comparison is the learning of the single trajectory of the rigid body dynamics contained in \S4.2.3 of that paper. The error of a single trajectory  (ground truth vs. a solution obtained by the neural network solution) is of the same order as our results. The error in Hamiltonian is somewhat better in our case, with the relative error being about $10^{-4}$ $(0.01\%)$ vs. $0.008$ in \citep{bajars2023locally}. However, that number depends on the particular realization of both neural networks and learning data, as we outlined above. The value of the Casimir $C(\bPi)=|\bPi|^2$ is conserved to machine precision in our case, whereas it would follow the general accuracy of computations in the previous work on the subject. 

\paragraph{Learning general dynamics of the rigid body} 
In the above calculation, we have followed the method of \citep{jin2022learning,bajars2023locally}
and learned the dynamics continuing a \emph{single} trajectory. However, it is also possible to extend the LPNets to learn several trajectories simultaneously, and predict the dynamics of the trajectory the method has not seen. We take the initial condition $\bar{\bPi}_0=(1/\sqrt{2},-1/\sqrt{2},1)$ and compute $20$ ground truth trajectories with the initial conditions $\bPi_0^j = \bar{\bPi}_0+ \epsilon_j$, where $\epsilon_j$ is a uniformly distributed random variable in the cube   $[-0.1,0.1] \times[-0.1,0.1]\times [-0.1,0.1] $. Each trajectory generates $200$ pairs of ground truth mapping between the momenta at the neighboring points, a total of $4000$ data points. 
A neural network is constructed with a similar structure as the one used for learning a  single trajectory,  having three inner layers with $32$ neurons, each having a sigmoid activation function, and the total of $2339$ trainable parameters. The neural network is trained using Adam algorithm with initial time step of $10^{-3}$ decaying exponentially to $10^{-4}$, over $10^5$ epochs. The final training and validation losses are slightly below $10^{-6}$ and $10^{-5}$ respectively, after 100,000 epochs. 

\begin{figure} 
\centering 
\includegraphics[height=0.26 \textheight]{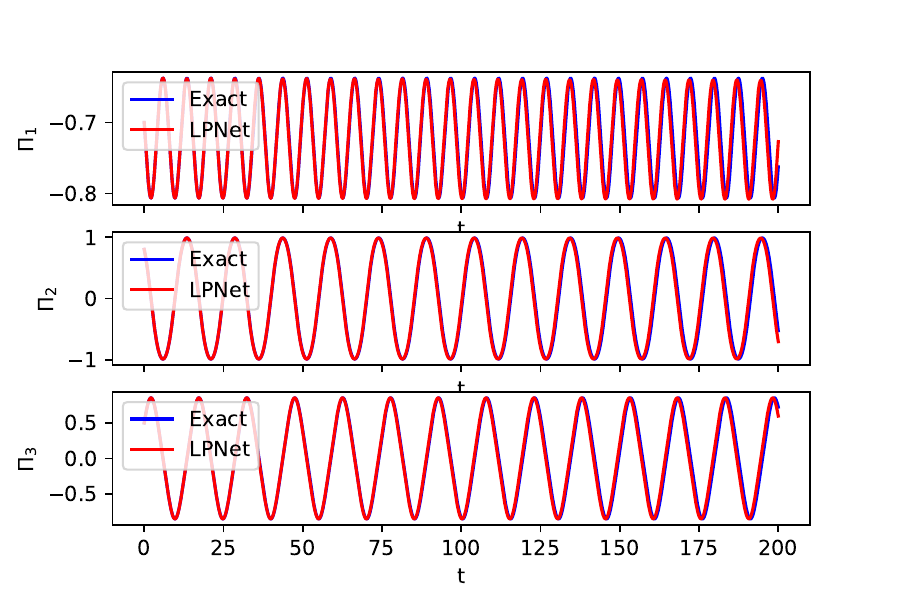}
\includegraphics[height=0.26 \textheight]{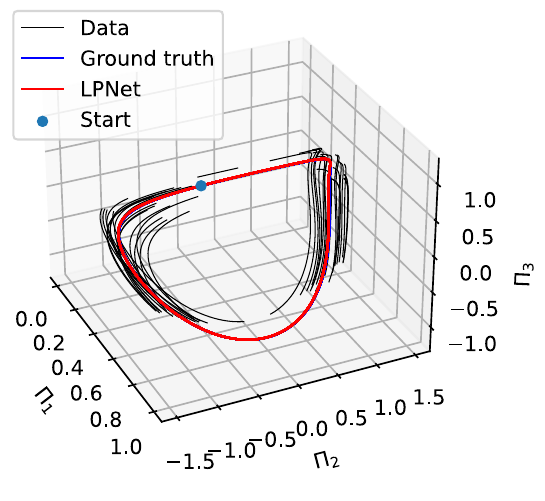}
\caption{Left: Same results as in Figure~\ref{fig:LP_results_rigid_body}, with the exception that the Neural Network is trained on several trajectories with random initial conditions different from the desired trajectory. Starting point is taken to be $\bPi_0=(1/\sqrt{2}, -1/\sqrt{2},1)$. 
Right: Trajectories in 3D space, similar to the presentation on the right panel of Figure~\ref{fig:LP_results_rigid_body}. In addition, the trajectories used for data learning are shown in black, and the blue dot indicates the starting point.  Trajectories from LPNets are shown in red, learning trajectories in black, and  trajectories considered ground truth are in blue.   \label{fig:LP_results_rigid_body_several_trajectories} }
\end{figure}

\begin{figure} 
\centering 
\includegraphics[width=0.48 \textwidth]{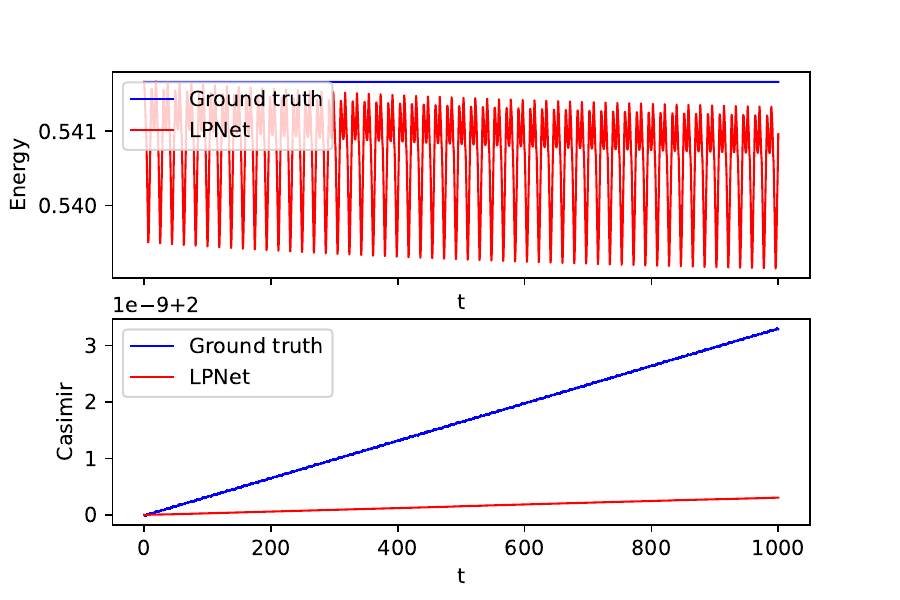}
\includegraphics[width=0.48 \textwidth]{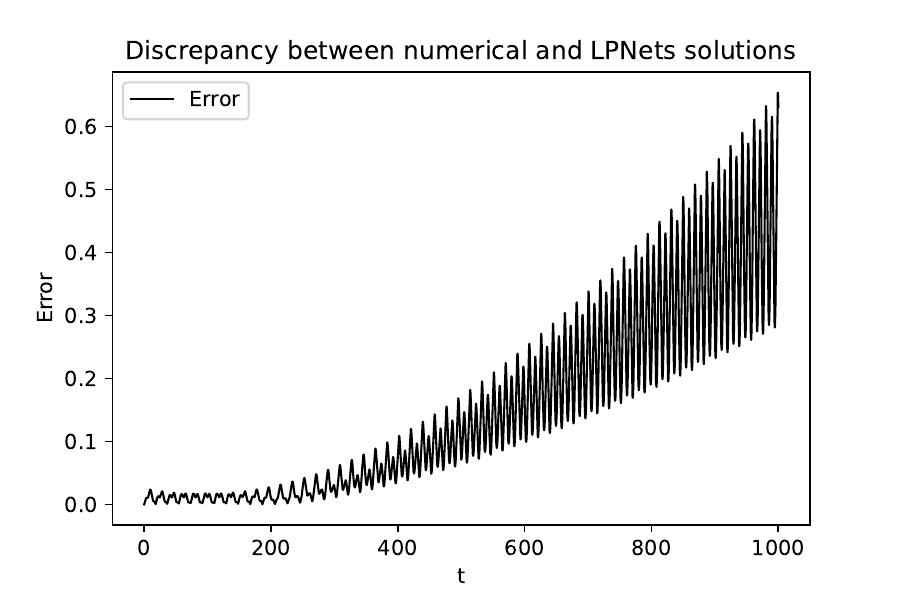}
\caption{Left: Conservation of the Hamiltonian $H$ (top) and the Casimir $C$ (bottom), comparing the results of LPNets (red) and ground truth (blue), for the case of global dynamics. Again, LPNets conserves the Casimir exactly (to machine precision) and thus substantially exceeds the ground truth in the conservation of Casimirs. Right: The discrepancy between the results of LPNets and the ground truth.  \label{fig:LP_Conservation_error_rigid_body_several_trajectories} }
\end{figure}

\paragraph{Trajectory prediction using LPNets} A trajectory is then constructed with the initial conditions $\bar \bPhi_0 = \bar{\bPi}_0$ iterating over $10000$ steps up to the time $t=1000$. Even though all the learning data were taken a finite distance from this solution, the LPNets faithfully reproduces the ground truth, as shown in Figure~\ref{fig:LP_results_rigid_body_several_trajectories}. While the solutions were computed until $t=1000$, the left panel of Figure~\ref{fig:LP_results_rigid_body_several_trajectories} only illustrates the evolution of momenta until $t=200$ for clarity. This example of $SO(3)$ shows that our method is capable of learning multiple trajectories and understanding the general Poisson dynamics. Of course, one has to keep in mind that in order to achieve good global accuracy, one needs to have quite a dense covering of the configuration manifold with the data points for learning, a task that may be difficult in many dimensions. A compromise presented here that is feasible to implement is to consider a few trajectories in the neighborhood of the desired trajectory for learning. To illustrate that point, we show the 3D plot of the trajectories and the corresponding data for learning on the right panel of Figure~\ref{fig:LP_results_rigid_body_several_trajectories}. Note that all the data were used on the right panel, and there is no visible deviation between the ground truth and the solutions in 3D.  We present the accuracy of the results compared to the ground truth and the preservation of the conserved quantities (Energy and Casimir) on Figure~\ref{fig:LP_Conservation_error_rigid_body_several_trajectories}. The conservation of the Hamiltonian is satisfied with the relative accuracy of about $0.25\%$ ($0.001$ in absolute accuracy), and the Casimir is conserved in our case to the absolute accuracy of about $10^{-11}$, several orders of magnitude better than the ground truth solution.

%\todo{VP: I re-run with much smaller amount of data ($4000$ instead of $100000$) and got similar results as before. The secret was to let the system iterate further and choose the best solution. I changed the numbers and the description in the above paragraph slightly. CE: Nice! }

\subsection{Extended pendulum case}

In order to compare our results directly with the Poisson Neural Networks developed in \citep{jin2022learning}, we consider the extended pendulum test case from that paper.
This example shows that our method extends beyond the Lie-Poissson case, as long as an explicit integration for the Poisson equations for linear Hamiltonians is available. This is case here since the Poisson tensor $\mathbb{B}( \mathbf{y})$ is affine in $ \mathbf{y}$.

Consider a standard pendulum of length $1$ and mass $1$, with the Hamiltonian 
$H = \frac{1}{2}p^2 - \cos q$. The equations of motion are written as 
\begin{equation}
\dot q = \pp{H}{p} = p \, , \quad \dot p = - \pp{H}{q} = - \sin q \,.
\label{pend_standard}
\end{equation}
The paper \citep{jin2022learning} then introduces an extra variable $c$ with equation $c=$const, extending the system to the three-dimensional space 
\begin{equation}
\frac{d}{dt} 
\left( 
\begin{array}{c} 
p
\\ 
q
\\ 
c
\end{array}
\right) 
= 
\left( 
\begin{array}{c} 
-\sin q 
\\ 
p+c 
\\ 
0 
\end{array}
\right) = 
\left( 
\begin{array}{ccc} 
0 & -1 & 0 
\\ 
1 & 0 &  0 
\\ 
0 & 0 & 0 
\end{array}
\right) \nabla_{(p,q,c)} \widetilde H \, , 
\label{pend_new}
\end{equation}
with the new Hamiltonian $\widetilde H = \frac{1}{2}p^2 - \cos q+ p c $. The paper \citep{jin2022learning} then 
 makes a transformation of variables 
\begin{equation}
\begin{aligned} 
(p,q,c) = \theta (u,v,r) = (u, v, r-u^2 -v^2) \, , 
\\
(u,v,r) = \theta^{-1} (p,q,c) = (p,q, p^2+q^2 +c) \, . 
\end{aligned}
\label{theta_def}
\end{equation} 
In the new variables $(u,v,r)$, the equations of motion become 
\begin{equation} 
\frac{d}{dt}
\left( 
\begin{array}{c} 
u 
\\ 
v
\\ 
r 
\end{array}
\right) 
= 
\left( 
\begin{array}{ccc} 
0 & -1 & -2 v  
\\ 
1 & 0 &  2 u  
\\ 
2 v & - 2 u & 0 
\end{array}
\right) \nabla_{(u,v,r)} K 
\label{ext_pendulum}
\end{equation} 
with the new Hamiltonian 
$K(u,v,r)=\frac{1}{2} u^2 - \cos v + u r - u^3 - uv^2$. The system \eqref{ext_pendulum} is Poisson with the corresponding bracket 
\begin{equation}
\{ F, H \} = (\nabla_\mathbf{y}  F)^T \cdot \mathbb{B}(\mathbf{y}) \cdot \nabla_\mathbf{y} H \, , \quad 
\mathbb{B}(\mathbf{y}):= \left( 
\begin{array}{ccc} 
0 & -1 & -2 v  
\\ 
1 & 0 &  2 u  
\\ 
2 v & - 2 u & 0 
\end{array}
\right) \,,
\label{bracket_ext_pendulum}
\end{equation} 
where we have denoted $\mathbf{y}= (u,v,r)^T$. The matrix $\mathbb{B}$ defined in \eqref{bracket_ext_pendulum} is degenerate, and 
\begin{equation} 
C(\mathbf{y})=r-u^2 - v^2 = y_3- y_1^2 - y_2^2\,,
\label{Casimir_pendulum} 
\end{equation} 
which is just $c$ in the old variables $(p,q,c)$, is a Casimir of the bracket \eqref{bracket_ext_pendulum}.  Indeed, one can readily check that $\mathbb{B}(\mathbf{y}) \cdot \nabla_{\mathbf{y}}C(\mathbf{y})=\mathbf{0}$. Moreover, one also checks that $\mathbb{B}(\mathbf{y})$ only has a single eigenvalue of $0$, so $C(\mathbf{y})$ defined in \eqref{Casimir_pendulum} is the only Casimir. 
In order to apply the method of LPNets, we take the test Hamiltonian linear in $\mathbf{y}$ as $H( \mathbf{y})=\boldsymbol{\alpha} \cdot \mathbf{y}$. For test Hamiltonians of that type, the equations of motion become 
\begin{equation} 
\dot{\mathbf{y}} = 
\left( 
\begin{array}{c} 
- \alpha_2 - 2 y_2 \alpha_3 
\\ 
\alpha_1 + 2 y_1 \alpha_3 
\\ 
2 \alpha_1 y_2 - 2 \alpha_2 y_1 
\end{array}
\right) \,.
\label{test_ham_eqs}
\end{equation} 
There are three test Hamiltonians to consider,  $H_a (\mathbf{y})= \alpha_a y_a$, $a=1,2,3$ (no sum). Then, the equations of motion are 
\begin{equation}
\left\{ 
\begin{aligned} 
H_1 = \alpha_1 y_1 & \Rightarrow \,  \dot y_1 =0 \, ,  \dot y_2 = \alpha_1 \, , \dot y_3 = 2 \alpha_1 y_2 
\\
H_2 = \alpha_2 y_2 & \Rightarrow \,  \dot y_1 =-\alpha_2 \, ,  \dot y_2 = 0 \, , \dot y_3 = - 2 \alpha_2 y_1
\\
H_3 = \alpha_3 y_3 & \Rightarrow \,  \dot y_1 =-2 y_2 \alpha_3 \, ,  \dot y_2 = 2 y_1 \alpha_3  \, , \dot y_3 = 0 \,.
\end{aligned}
\right. 
\label{H_test_equations}
\end{equation} 
Equations \eqref{H_test_equations} are easily solved explicitly. Each Hamiltonian $H_a= \alpha_a y_a$ leads to explicit expressions for an affine transformation $\mathbf{T}(t, \alpha_a, \mathbf{y}_0)$ of the initial condition $\mathbf{y}_0$ to the final solution after time $t$: 
\begin{equation} 
H_a= \alpha_a y_a \,   \Rightarrow \,  \mathbf{y}= \mathbf{T}_a(t, \alpha_a,\mathbf{y}_0) 
\label{H_test_solutions}
\end{equation}
with the transformations $\mathbf{T}_a$ given by 
\begin{equation}
\begin{aligned} 
\mathbf{T}_1 (t, \alpha_1,\mathbf{y}_0) & = 
\left( 
\begin{array}{c} 
y_1(0) 
\\ 
y_2(0) + \alpha_1 t
\\
 y_3(0) + 2 t \alpha_1 y_2(0) + \alpha_1^2 t^2 
\end{array}
\right) 
\\
\mathbf{T}_2 (t, \alpha_2,\mathbf{y}_0) & = 
\left( 
\begin{array}{c} 
y_1(0)-\alpha_2 t
\\ 
y_2(0) 
\\
 y_3(0) - 2 t \alpha_2 y_1(0) + \alpha_2^2 t^2 
\end{array}
\right) 
\\
\mathbf{T}_3 (t, \alpha_3,\mathbf{y}_0) &  = 
\left( 
\begin{array}{c} 
y_1(0) \cos (2 \alpha_3 t ) -y_2(0) \sin (2 \alpha_3 t ) 
\\ 
y_1(0) \sin (2 \alpha_3 t ) +y_2(0) \cos (2 \alpha_3 t ) 
\\
y_3(0) 
\end{array}
\right) \,.
\end{aligned} 
\label{T_i_operators}
\end{equation} 

\paragraph{Data preparation, Part 1: finding the transformations} Suppose we have pairs of solution $(\mathbf{y}_i, \mathbf{y}_{i+1})$ that are obtained from snapshots of a single or several trajectories of equation \eqref{ext_pendulum}; the time difference between the snapshots is $\Delta t=h$. We separate the interval $\Delta t=h$ between the snapshots into three equal sub-intervals (although other divisions of the time interval are also possible). 
Then, for each initial point of the pair $\mathbf{y}_i$ we are looking for the sequence of parameters $(\alpha_1, \alpha_2, \alpha_3)$ defining the transformations $\mathbf{T}_i$ as in \eqref{T_i_operators}, such that 
\begin{equation} 
\begin{aligned} 
\mathbf{y}_i^1 & = \mathbf{T}_1 \left( \frac{h}{3}, \alpha_1, \mathbf{y}_i \right) \, , 
\\
\mathbf{y}_i^2 & = \mathbf{T}_2 \left( \frac{h}{3}, \alpha_2, \mathbf{y}_i^1 \right) \, , 
\\
\mathbf{y}_i^3 & = \mathbf{T}_3 \left( \frac{h}{3}, \alpha_3, \mathbf{y}_i^2 \right) = \mathbf{y}_{i+1}\, . 
\end{aligned}
\label{alpha_def}
\end{equation} 
In other words, $\overline{\boldsymbol{\alpha}}_i = (\alpha_1, \alpha _2, \alpha _3)(\mathbf{y}_i)$ should be such that the three transformations $\mathbf{T}_{1,2,3}$, performed in the corresponding order, map the initial snapshot of the pair $\mathbf{y}_i$ to the final snapshot of the pair $\mathbf{y}_{i+1}$. 

For data that is not precise, the matching of $\overline{\boldsymbol{\alpha}}_i$ to the data can be accomplished numerically using a gradient descent method, with further removal of components of the Casimir gradient $\nabla C$. Indeed, 
each set of parameters $\overline{\boldsymbol{\alpha}}_i = (\alpha_1,\alpha_2, \alpha_3)(\mathbf{y}_i)$ determined in the previous step, is only defined up to the value of $\nabla C$. Since $\nabla C$ is a zero eigenvector of the matrix $\mathbb{B}(\mathbf{y})$, changing \begin{equation} 
\boldsymbol{\alpha}_i \rightarrow \boldsymbol{\alpha}_i + k \nabla C 
\label{alpha_shift}
\end{equation} 
for arbitrary $k$ does not change the result of composition of transformations $\mathbf{T}_3 \circ \mathbf{T}_2 \circ \mathbf{T}_1$. We thus choose $k$ in \eqref{alpha_shift} such that the projection of $\boldsymbol{\alpha}$ on $\nabla C$ vanishes, \emph{i.e.}, take 
\begin{equation} 
\boldsymbol{\alpha}_i^* =\boldsymbol{\alpha}_i -  \nabla C
\frac{\boldsymbol{\alpha}_i \cdot \nabla C}{|\nabla C |^2} \,.
\label{alpha_shift_calc}
\end{equation} 
However, in the case the training data is \emph{exact}, for example, is obtained from a numerical simulation with a very high accuracy, we can solve equations for $\overline{\boldsymbol{\alpha}}_i$ analytically. For shortness, we denote the components of $\mathbf{y}_i$ as $(y_1,y_2,y_3)$, dropping the index $i$, and the components of $\mathbf{y}_{i+1}$  by $(y_{1,f},y_{2,f},y_{3,f})$, since $\mathbf{y}_{i+1}$ represents the final value of the interval $t \in (t_i,t_{i+1})$. Let us notice that the sequential application of $\mathbf{T}_1$ and $\mathbf{T}_2$ after time $\Delta t=h/3$ on each sub-step gives the intermediate values 
\begin{equation}
\begin{aligned} 
y_1 &  \rightarrow y_1^*=y_1 - \alpha_2 \Delta t \, , 
\\
y_2 &\rightarrow y_2^*=y_2 + \alpha_1 \Delta t \, , 
\\
y_3 & \rightarrow y_3^* =y_3  + (\alpha_1-\alpha_2) \Delta t + ( \alpha_1^2 + \alpha_2^2) \Delta t^2  \, .
\end{aligned} 
\label{y_12_intermediate}
\end{equation} 
On the third sub-step, $y_3$ doesn't change so $y_3^*=y_{3,f}$. Also, at that sub-step, the application of $\mathbf{T}_3$ just induces a rotation of $(y_1^*,y_2^*)$ by the angle $2 \alpha_3 \Delta t$, so the compatibility conditions to match the data points precisely is 
\begin{equation} 
y_3^* = y_{3,f}\, , \quad 
(y_1^*)^2+(y_2^*)^2= y_{1,f}^2 +y_{2,f}^2 \, . 
\label{cond_third_step}
\end{equation}
Using \eqref{y_12_intermediate}, we can rewrite \eqref{cond_third_step} as 
\begin{equation} 
y_{3,f} - y_{1,f}^2 - y_{2,f}^2 = y_{3} - y_{1}^2 - y_{2}^2  \, , 
\label{Casimir_cons_pendulum_step}
\end{equation} 
so the solution for $\overline{ \boldsymbol{\alpha}}_i$ giving exact match between the data points can be found if and only if the Casimir is exactly the same for all values of the data. If the data contains noise, an optimization procedure must be introduced searching for an optimal value of $\overline{ \boldsymbol{\alpha}}_i$ at every step. In this paper, we assume that the learning data is exact. In that case, we can set $\alpha_3=0$ on every time step and match the values of the components $y_1$ and $y_2$; the matching of the component $y_3$ is done automatically due to the conservation of the Casimir. 
\begin{equation}
\alpha_1 = \frac{3}{h} \left( y_{2,i+1}-y_{2,i} \right) 
\, , \quad 
\alpha_2 = -\frac{3}{h} \left( y_{1,i+1}-y_{1,i} \right) 
\, , \quad 
\alpha_3 =0 \, . 
\label{alpha_sol_pendulum}
\end{equation}

\paragraph{Learning procedure: Neural Network approximation}
The inputs $\mathbf{y}_i$ and outputs $\overline{ \boldsymbol{\alpha}}_i$ computed from \eqref{alpha_sol_pendulum} are used to learn the mapping $\overline{\boldsymbol{\alpha}}(\mathbf{y})$ using a Neural Network. 
In order to match \citep{jin2022learning}, we use three trajectories with initial conditions $\mathbf{y}_0 = (0,1,1^2)$, $(0,1.5,1.5^2+0.1)$ and $(0,2,2^2+0.2)$, with the time step of $h=0.1$. We use the number of points $N=1000$ with half of these points used for training and half for validation. The ground truth solution is computed using Python's \emph{Scipy odeint} routine with \emph{BDF} (Backward Differentiation Formula), with the tolerances, relative and absolute, set at $10^{-13}$ and $10^{-14}$, respectively. 

\begin{figure}
\centering
\includegraphics[height=0.26 \textheight]{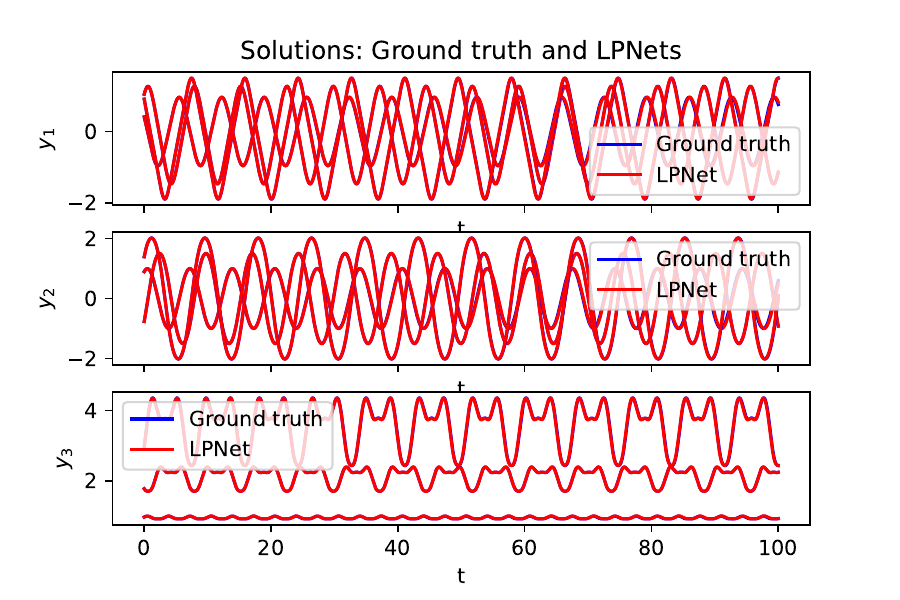}
\includegraphics[height=0.26 \textheight]{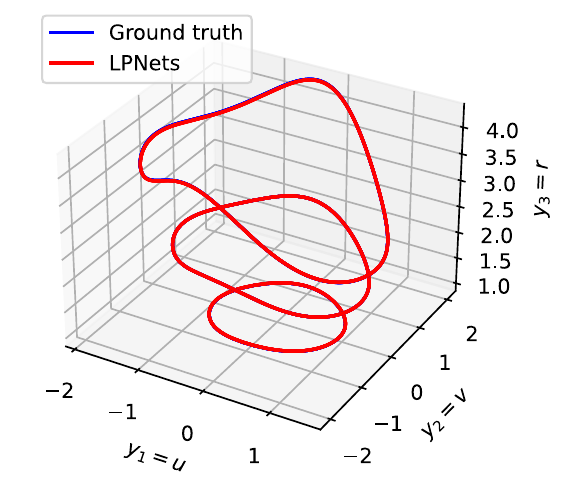}
\caption{Results of simulation of equations \eqref{ext_pendulum} and the corresponding solution of  using LPNets procedure. Left panel: three components of the solution $y_1=u$, $y_2=v$ and $y_3=r$ versus time. Right panel: phase space plot of the solution. Blue line: high-precision numerics taken as the exact solution; red line: LPNets. }
\label{fig:sim_pendulum}
\end{figure}

\begin{figure}
    \centering
    \includegraphics[width=0.45 \textwidth]{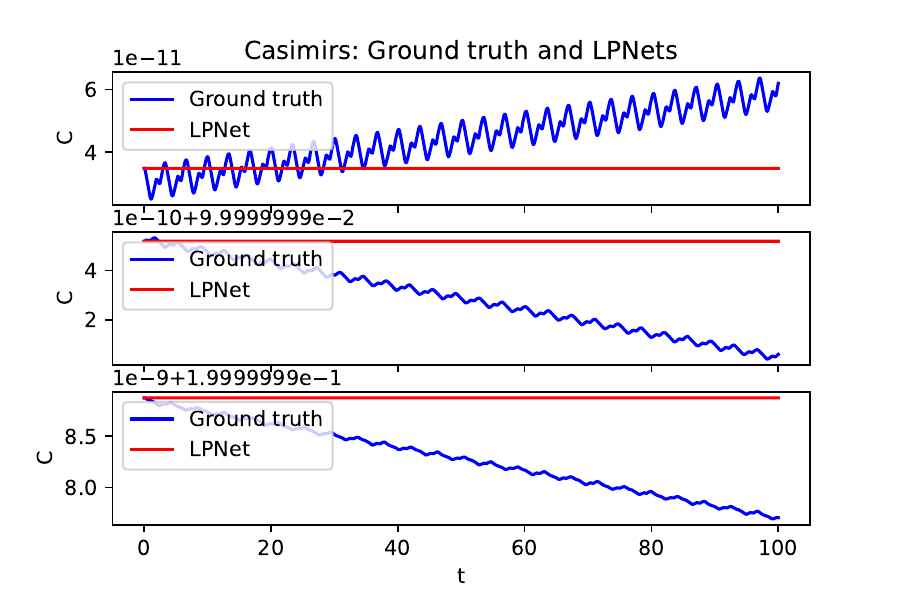}
    \includegraphics[width=0.45 \textwidth]{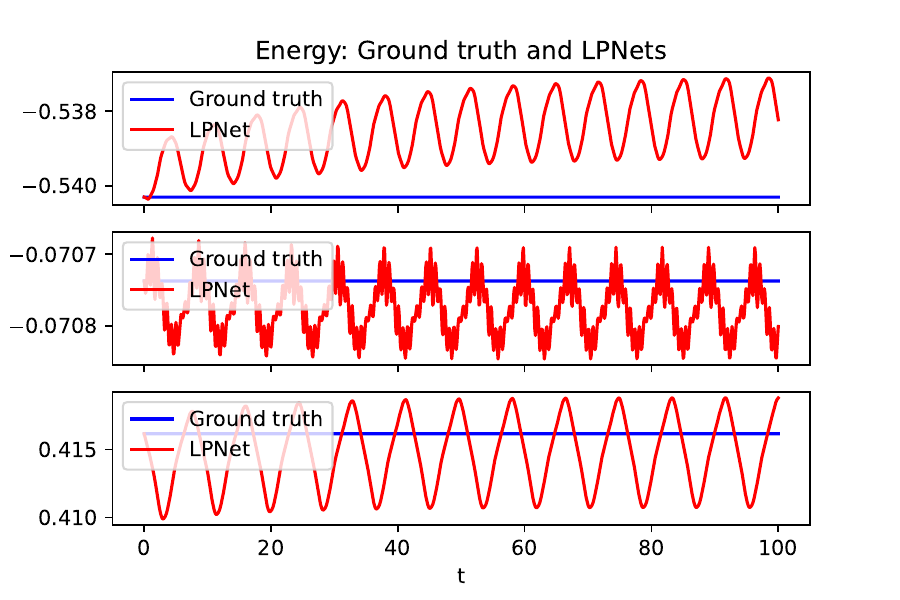}
    \caption{Left: Conservation of the Casimir $C=r-u^2-v^2$ in the solutions of equations \eqref{ext_pendulum} obtained by the high precision numerics (blue line) and the corresponding solution of  using LPNets procedure (red line). Even though the precision of numerics is $10^{-11}$, LPNets is substantially more precise, as it achieves machine precision of Casimir conservation on each time step. Right: Conservation of the energy of the system for all three cases. The relative accuracy in the conservation of energy is about $0.5-1$\%. }
    \label{fig:Casimir_pendulum}
\end{figure}

\paragraph{Evaluation and dynamics using Neural Network}
Just as in \citep{jin2022learning}, we compute the trajectories starting at the last point of the trajectory $\mathbf{y}_N$ for $1000$ points, with half of these points used for training and half for validation. We use a Sequential Keras program, with three inner layers of neurons with sigmoid activation functions. Each layer has the breadth of $16$ neurons with the total number of tunable parameters being $642$. The network uses the Adam optimizer with the learning rate of $10^{-3}$, exponentially decaying to $10^{-5}$ and Mean Square Error loss, computed for $50000$ epochs. The resulting MSE achieved is between $10^{-7}$ and $10^{-8}$. The results of the simulations are shown in Figure~\ref{fig:sim_pendulum}. On the left side of this Figure, we show the match between the "exact" solution and the solution obtained by LPNets (notice that the "ground truth" solution is still obtained using a numerical method with a given accuracy).  On the right hand side of that figure, we present a three-dimensional plot of the phase space $(u,v,r)=(y_1,y_2,y_3)$ comparing the numerical results in blue with the LPNets results in red. 
In Figure~\ref{fig:Casimir_pendulum}, we show the conservation of the Casimir for all three initial conditions. Note that the ground truth numerical solution (blue line) only conserves the Casimir $C=r-u^2-v^2$ up to the accuracy of calculation, accumulating the error of the order $10^{-9}$ after $t=100$. In contrast, the Casimir in LPNets (red line) is preserved to machine precision by the very nature of transformations performed on each time step. On the right panel of that Figure, we present the accuracy of the evolution of Energy in LPNets. As one can see, the energy conservation by LPNets is quite satisfactory, yielding the relative error of about $1\%$ or less for all cases. 

\begin{figure}
    \centering
    \includegraphics[width=0.95 \textwidth]{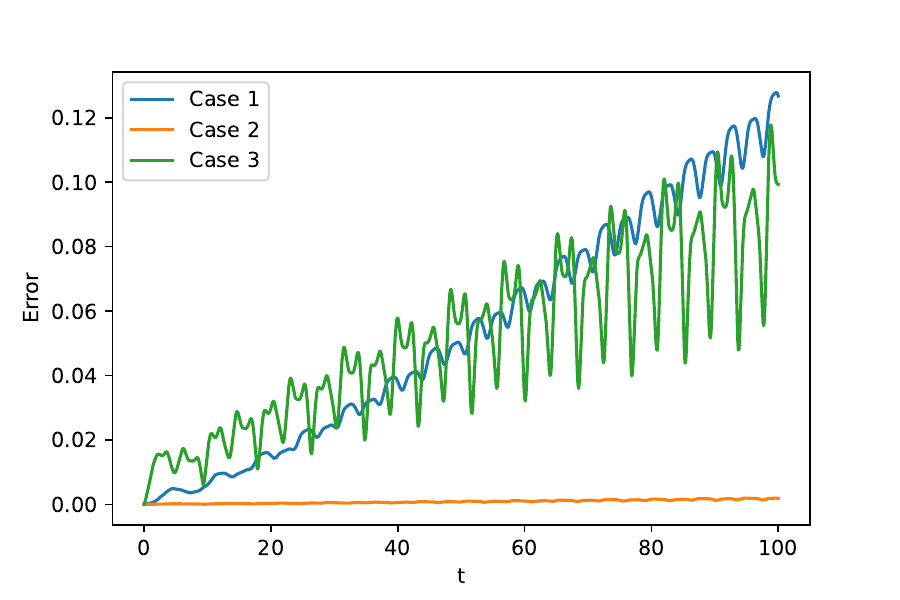}
    \caption{The discrepancy between the ground truth and the solutions provided by LPNets for the extended pendulum case.  }
    \label{fig:Error_pendulum}
\end{figure}

Finally, in Figure~\ref{fig:Error_pendulum}, we present the results for the discrepancy between the ground truth and LPNets solutions. Even though the discrepancy grows, it is mostly due to the fact that the time between the ground truth and the solution is slightly mismatched, which explains why energy is conserved to much higher precision than the solution itself. The accuracy is still quite good and the solution obtained by the LPNets is virtually indistinguishable from the ground truth on the left side of the Figure~\ref{fig:sim_pendulum}. 

Finally, we would like to emphasize that a system of three inner layers having a width of 16 in each layer should be viewed as very compact. Higher accuracy can be achieved with more data points and correspondingly wider or deeper network, or having more insights into the structure of the framework which we have assumed to be completely dense. Achieving these efficiencies is an interesting challenge that we will consider in the future. 

\paragraph{Comparison with previous results for extended pendulum system}
The visual agreement of the solutions and the ground truth is similar to the one presented in \citep{jin2022learning} for this system. The conservation of the Casimir is not presented in \citep{jin2022learning}. We interrogated the data produced by the code made available in \citep{jin2022learning} and found that the relative errors for Casimir in the transformed Lie-Darboux coordinates are quite small, between $10^{-7}$ and $10^{-6}$.
The errors in Casimir in the original coordinates $(u,v,r)$ for the parameters presented in \citep{jin2022learning}, although still quite small, are several orders of magnitude larger than the transformed coordinates. 
This could be attributed to the fact that the inverse of Lie-Darboux transformation is not being computed sufficiently accurately. This accuracy of PNNs can certainly be improved by appropriate modification of the original PNN network from \citep{jin2022learning}, which is beyond the scope of the paper. In any case, our method conserves the Casimir exactly for the original system with no necessity to search for the Lie-Darboux transformation to the canonical coordinates. 

\subsection{A particle in a magnetic field}
To compare with the second test case computed in \citep{jin2022learning}, we study a particle of mass $m$ and charge $q$ moving in a   magnetic field $\mathbf{B}( \mathbf{x} )$. We assume that the motion of the particle is in $\mathbf{x} \in \mathbb{R}^3$, and the relevant variables are the particle position $\mathbf{x}=(x_1,x_2,x_3)$ and its' momentum $\mathbf{p} = (p_1, p_2, p_3)$. The equations of motion for a particle in a magnetic field $\mathbf{B}(\mathbf{x})$ are: 
\begin{equation}
\frac{d}{dt} 
\left( 
\begin{array}{c} 
\mathbf{p} 
\\ 
\mathbf{x} 
\end{array}
\right) 
= 
\left( 
\begin{array}{cc} 
- \frac{q}{m} \widehat{B} ( \mathbf{x} ) & -   \mathbb{I}_3\\ 
\mathbb{I}_3 & 0 
\end{array} 
\right) 
\left( 
\begin{array}{c} 
\pp{H}{\mathbf{p}}
\\ 
\pp{H}{\mathbf{x}}
\end{array}
\right) := \mathbb{B} ( \mathbf{x})\nabla_{(\mathbf{p}, \mathbf{x})} H 
\label{charged_particle_eq}
\end{equation} 
Notice that equations \eqref{charged_particle_eq} are not of Lie-Poisson form. However, explicit solutions for the transformations on each time step can be found here as well. We thus believe that this problem is a useful case for demonstrating the power and applicability of these methods beyond the Lie-Poisson equations. 

The Hamiltonian for simulations is taken as 
\begin{equation}
H( \mathbf{x} , \mathbf{p} )= \frac{1}{2m} | \mathbf{p}|^2 + q \varphi(\mathbf{x}) \, . 
\label{Hamiltonian_charged_particle}
\end{equation}
Here, $\mathbb{I}_3$ is, as usual, a $3\times 3$ unity matrix, and we used the hat map notation 
\begin{equation} 
\widehat{B}(\mathbf{x}) = 
\left( 
\begin{array}{ccc} 
0 & -B_3(\mathbf{x})  & B_2(\mathbf{x})
\\ 
B_3(\mathbf{x}) & 0  & -B_1(\mathbf{x})
\\ 
 -B_2(\mathbf{x}) & B_1(\mathbf{x})  & 0
\end{array} 
\right) \, . 
\label{hatmap_B}
\end{equation} 
Note that $\widehat{B}( \mathbf{x} ) \mathbf{v} = \mathbf{B}( \mathbf{x} ) \times \mathbf{v}$  for all $ \mathbf{x} \in \mathbb{R}^3$. Similar to \citep{jin2022learning,bajars2023locally}, we take the following values for parameters,  electric potential $\varphi(\mathbf{x})$ and the magnetic field $\mathbf{B}(\mathbf{x})$: 
\begin{equation} 
\begin{aligned} 
& q=1, \quad   m=1 , 
\\ 
& \mathbf{B} = (0, 0, B_3) \, ,  \quad \mbox{with}  \quad B_3(\mathbf{x}) = \sqrt{x_1^2+x_2^2}\,,
\\ 
& \varphi(\mathbf{x}) = \frac{1}{100  \sqrt{x_1^2+x_2^2} } \,.
\end{aligned} 
\label{parameters_charged_particle}
\end{equation} 
One can readily check that \eqref{charged_particle_eq} possesses no Casimirs since $\mathbb{B}$ is non-degenerate.
%. Indeed, for Casimirs to exist, we need to find $\mathbb{B} (\mathbf{v},\mathbf{x})^T=\mathbf{0}$. From the $\mathbf{x}$-part of the equation, we get $\mathbf{v}=0$, and hence from $\mathbf{v}$-part of the equation gives $\mathbf{x}=0$. Thus, $\mathbb{B}$ is non-degenerate and there are no Casimirs. 

\paragraph{Reduction of motion and conserved quantities}
In \citep{jin2022learning}, the initial conditions were chosen to be $(x_3=0, v_3=0)$ so the particle would always move on the plane. The motion in that case is four-dimensional, and we only need four test Hamiltonians. We thus take the Hamiltonians linear in velocities $(v_1,v_2)$ and coordinates $(x_1,x_2)$ and compute the corresponding motion. 

Note that when the system \eqref{charged_particle_eq} is restricted so both $\mathbf{x}$ and $\mathbf{p}$ are in the plane, \emph{i.e.} both $x_3=0$ and $p_3=0$, and for the choice of any $\mathbf{B} = \mathbf{e}_3 B_3(r)$ and $\varphi = \varphi(r)$, where $r = \sqrt{x_1^2+x_2^2}$, there are two integrals of motion. One is clearly the Hamiltonian \eqref{Hamiltonian_charged_particle}. Another one can be found by considering the evolution for the angular momentum in the   $\mathbf{e}_3$ direction $M_3 = \mathbf{e}_3 \cdot (\mathbf{x} \times \mathbf{p})$. We can observe that 
\begin{equation} 
\dot M_3 = - q B_3 (r) \mathbf{x} \cdot \dot{\mathbf{x}} = - q B_3 r \dot r 
\label{generalized_momentum_evolution}
\end{equation}
leading to the conservation law that for \eqref{parameters_charged_particle} and $q=1$ reduces to 
\begin{equation}
I = M_3 +q  \int^r B_3(s) s \mbox{d} s = x_1 p_2 - x_2 p_1 + \frac{1}{3} \left( x_1^2+x_2^2 \right)^{3/2} = \rm{const}. 
\label{Cons_law_momentum}
\end{equation}
Therefore, the system \eqref{charged_particle_eq}, for the choice of functions \eqref{parameters_charged_particle} and reduced to four-dimensional motion, is essentially a two-dimensional motion because of the conservation of the Hamiltonian \eqref{Hamiltonian_charged_particle} and \eqref{Cons_law_momentum}. Thus, one should expect a limited richness of solution behavior. Nevertheless, it is a good test problem and since it has been used in both recent papers on the subject \citep{jin2022learning,bajars2023locally}, we study this particular case as well. 

\paragraph{Lie-Poisson transformations}
Suppose we have a set of $N$ data pairs, each pair is obtained by the phase flow from $\mathbf{y}_i = (\mathbf{x}_i,\mathbf{p}_i)$ to some value $\mathbf{y}_i^f = (\mathbf{x}_i^f, \mathbf{p}_i^f)$. If these pairs are obtained from a single trajectory, then $\mathbf{y}_i^f = \mathbf{y}_{i+1}$, although this does not have to be the case -- our method is capable of learning from several trajectories. 
In order to apply our method, we need to compute the results of phase flows for Hamiltonians linear in coordinates $\mathbf{x}$ and momenta $\mathbf{p}$. 

We just present the answers for these transformations here for brevity; an interested reader may readily check these formulas.  In what follows, we shall use the function $\boldsymbol{\Phi}$ of time $t$, initial conditions $\mathbf{X}=\mathbf{x}(0)$ and parameters $ \boldsymbol{\alpha}=(\alpha_1,\alpha_2)$ defined as follows: 
\begin{equation} 
\boldsymbol{\Phi} (t; \mathbf{X}, \boldsymbol{\alpha}) = 
\int^t_0 B_3 ( \mathbf{X} + \boldsymbol{\alpha}  s )\mbox{d} s \,.
\label{gen_Phi}
\end{equation}
For the expression of $B_3(\mathbf{x})$ taken in \eqref{parameters_charged_particle},  both components of the function $ \boldsymbol{\Phi}$ can be expressed in terms of a single elementary function coming from taking the quadrature of \eqref{gen_Phi}:
\begin{equation} 
\begin{aligned} 
\Phi (t; \mathbf{X},  \boldsymbol{\alpha}) & = 
\frac{1}{2 A^3} \left( A (X_1\alpha_1 +X_2 \alpha_2 +A^2 t) \sqrt{ x_1(t)^2+x_2(t)^2 } \right. 
\\ 
&  \qquad + (X_1 \alpha_2 - X_2 \alpha_1)^2 \\
& \qquad \left. \times \log \left[ A \sqrt{x_1(t)^2+x_2(t)^2} + (X_1\alpha_1 +X_2 \alpha_2 +A^2 t) \right]\right) \, , 
\\ 
& A:= | \boldsymbol{\alpha}| = \sqrt{\alpha_1^2+\alpha_2^2}\, ,
\quad x_i(t) := X_i + \alpha_i t, \, i=1,2 \,. 
\label{integrated_Phi}
\end{aligned}
\end{equation}
The presence of explicit expression for the function \eqref{gen_Phi} is helpful, but it is not essential; for general $B_3(\mathbf{x})$, one can use the quadrature expressions \eqref{gen_Phi}. The transformations for each test Hamiltonian are: 
\begin{enumerate}
\item $H_1 = \alpha_1 p_1+\alpha_2 p_2$ leads to the motion $\mathbf{y}=\mathbf{T}_1(t,\mathbf{p}_0, \mathbf{x}_0)$
\begin{equation} 
\left( 
\begin{array}{c}
p_1 (t) 
\\
p_2(t) 
\\
x_1(t) 
\\ 
x_2(t) 
\end{array}
\right) = 
\mathbf{T}_1(t,\mathbf{p}_0,  \mathbf{x}_0) = 
\left( 
\begin{array}{l}
-\Phi(t;  \mathbf{x}_0,\boldsymbol{\alpha}) + \Phi(0; \mathbf{x}_0,\boldsymbol{\alpha})+ p_1(0)
\\ 
\Phi(t;  \mathbf{x}_0,\boldsymbol{\alpha}) - \Phi(0; \mathbf{x}_0,\boldsymbol{\alpha}) + p_2(0) 
\\ 
\alpha_1 t + x_1(0) 
\\ 
\alpha_2 t + x_1(0) 
\end{array}
\right) \, , 
\label{T1_charged_particle}
\end{equation} 
\item $H_2 = \beta_1 x_1+\beta_2 x_2$ leads to the motion $\mathbf{y}=\mathbf{T}_2(t,\mathbf{p}_0, \mathbf{x}_0)$
\begin{equation} 
\left( 
\begin{array}{c}
p_1 (t) 
\\
p_2(t) 
\\
x_1(t) 
\\ 
x_2(t) 
\end{array}
\right) = 
\mathbf{T}_2(t,\mathbf{p}_0, \mathbf{x}_0) = 
\left( 
\begin{array}{l}
- \beta_1 t  + p_1(0) 
\\
- \beta_2 t  + p_2(0) 
\\ 
x_1(0) 
\\
x_2(0) 
\end{array}
\right) \, . 
\label{T2_charged_particle}
\end{equation} 
\end{enumerate} 

Clearly, $\mathbf{T}_{2}$ does not alter the coordinates $x_{1,2}$. Thus, the algorithm of computation is very explicit, as $\alpha_{1,2}$ and $\beta_{1,2}$ can be solved without any root-finding procedure in the learning stage. 
\paragraph{Data preparation}
Divide the interval $\Delta t = h$ into two equal steps $h/2$. On each of those steps, for each pair of data points $(\mathbf{y}_i, \mathbf{y}_{i}^f)$ where $\mathbf{y} = (p_1,p_2,x_1,x_2)$, we compute the parameters $(\alpha_{1,i},\alpha_{2,i},\beta_{1,i},\beta_{2,i})$ as follows:
\begin{enumerate} 
\item Compute  $\alpha_{1,2}$ at the $i$-th data point to match $x_{1,2}$, and corresponding new intermediate momenta $p_{1,2}^*$ by using the corresponding transformations $\mathbf{T}_{1,2}$ according to 
\begin{equation} 
\begin{aligned} 
 \alpha_{1,i} & = \frac{2}{h} \left( x_{1,i}^f - x_{1,i} \right)
\\ 
 \alpha_{2,i} & = \frac{2}{h} \left( x_{2,i}^f - x_{2,i} \right)
\\ 
 p_{1,i}^* & = p_1 -\Phi(h/2; \mathbf{x}_0,\boldsymbol{\alpha}) + \Phi(0; \mathbf{x}_0,\boldsymbol{\alpha}) 
\\ 
 p_{2,i}^* & = p_2+\Phi(h/2; \mathbf{x}_0,\boldsymbol{\alpha}) - \Phi(0; \mathbf{x}_0,\boldsymbol{\alpha})\,.
\end{aligned} 
\end{equation} 
\item Compute $\beta_{1,2}$ at the $i$-th data point according to 
\begin{equation} 
\begin{aligned} 
\beta_{1,i} & = \frac{2}{h} \left(p_{1,i}^*-p_{1,i}^f \right) 
\\ 
\beta_{2,i} & = \frac{2}{h} \left(p_{2,i}^*-p_{2,i}^f \right) \,.
\end{aligned} 
\end{equation} 
\end{enumerate} 
\paragraph{Learning using Neural Network}
The network will learn the mapping between $\mathbf{y}$ and the parameters $(\alpha_1,\alpha_2, \beta_1, \beta_2)$ using $\mathbf{y}_i$ as inputs and corresponding parameters $(\boldsymbol{\alpha}_i,\boldsymbol{\beta}_i)$ as outputs. 

The network structure consists of an input layer taking four inputs: $(\mathbf{x}, \mathbf{p})$, three densely connected hidden layers of $32$ neurons,  with the sigmoid activation function, and the output layer producing an estimate for four outputs $(\boldsymbol{\alpha},\boldsymbol{\beta})$. The number of trainable parameters in the network is $2404$. Adam optimization algorithm uses the step $10^{-3}$ decaying exponentially to $10^{-5}$, with the number of epochs equal to $2\cdot 10^5$. The loss function is taken to be the mean square average (MSE).  Because of the large number of parameters, to avoid overfitting, we take $20000$ data points of a single trajectory separated by the time step $h=0.1$.  For that trajectory, use the initial conditions $\mathbf{v}_0= (1,0.5)$ and $\mathbf{x}_0=(0.5,1)$. From all the data points, $80\%$ are used for learning and $20\%$ for evaluation. At the end of the learning procedure, both the loss and the validation loss drop to values below $10^{-5}$. 
 
\begin{figure} 
\centering 
\includegraphics[width=1\textwidth]{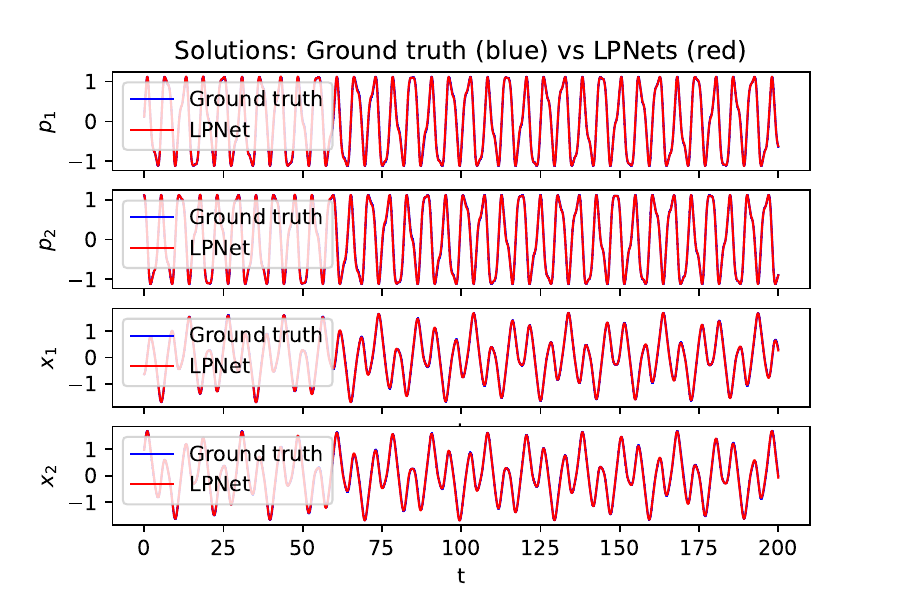}
\caption{Solutions of \eqref{charged_particle_eq}: ground truth (blue lines) vs an iterative solution obtained by LPNets (red lines).  }
\label{fig:particle_solutions}
\end{figure}

\paragraph{Solution evaluations}
We choose the initial condition to coincide with the end of the learning trajectory, which is located at $\mathbf{x}_0 \simeq (-0.627,  0.985)^T$ and $\mathbf{p}_0 \simeq (0.119,  1.112)^T$. We present the results of simulations in Figure~\ref{fig:particle_solutions}. The ground truth was obtained by solving \eqref{charged_particle_eq} numerically with the given initial conditions using the BDF solver of \emph{SciPy}, with the relative tolerance of $10^{-13}$ and absolute tolerance $10^{-14}$ for $t=200$, providing outputs every $h=0.1$.  The ground truth is presented with a solid blue line. 

\begin{figure} 
\centering 
\includegraphics[width=0.45\textwidth]{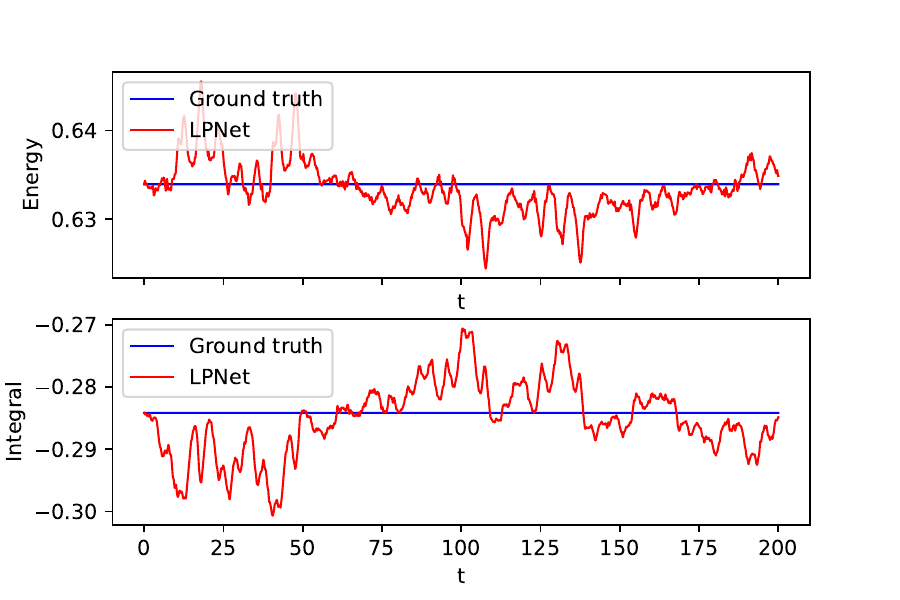}
\includegraphics[width=0.45\textwidth]{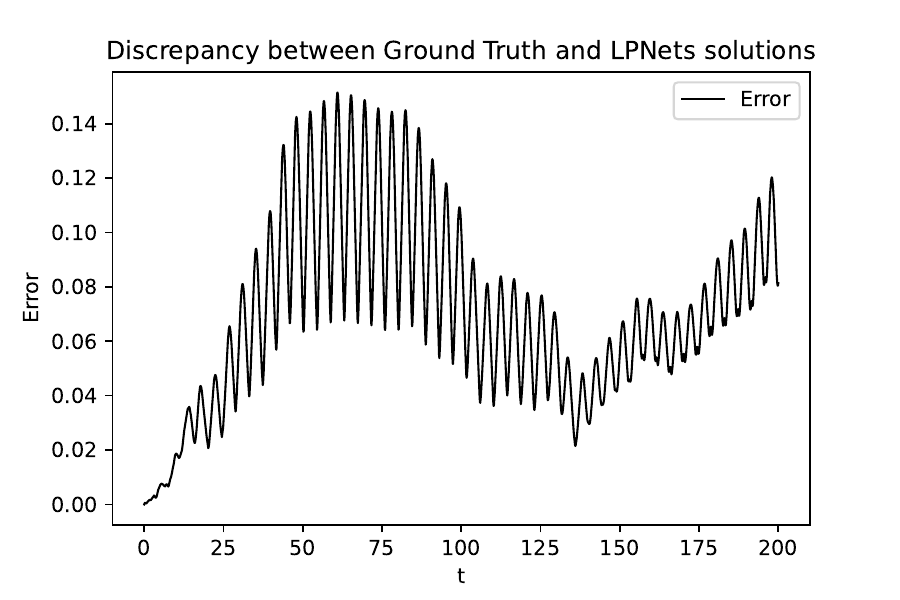}
\caption{Left: the value of Hamiltonian (energy) obtained from \eqref{Hamiltonian_charged_particle} and the conserved quantity \eqref{Cons_law_momentum}. The  ground truth is plotted as blue lines, and the results from the iterative solution obtained by LPNets are shown as red lines.  Right: mean-square discrepancy (in all components) between the ground truth and the results obtained by the LPNets.  }
\label{fig:particle_error}
\end{figure}

The iterative solution using the evaluation provided by successful iterations of Poisson maps \eqref{T1_charged_particle} and \eqref{T2_charged_particle} with the parameters $(\boldsymbol{\alpha},\boldsymbol{\beta})$ provided by the neural network, approximates the solution at the time points $t=i h$, $i=1, \ldots 2000$. The iteration starts with the same initial conditions as the ground truth solutions. The results of the iterations are presented in Figure ~\ref{fig:particle_solutions} with solid red lines. As one can see, the LPNets solution approximates ground truth quite well. 
To further investigate the results, on the left panel Figure~\ref{fig:particle_error} we show the mean-square deviation between the ground truth and LPNets solutions of \eqref{charged_particle_eq} presented in Figure~\ref{fig:particle_solutions}. On the right panel of this Figure, we present the conserved quantities: Hamiltonian (Energy) $H$ given by \eqref{Hamiltonian_charged_particle} and the conserved quantity \eqref{Cons_law_momentum}. As one can see, the Hamiltonian is preserved with the relative accuracy of $1-2\%$ and the integral \eqref{Cons_law_momentum} to about $3-4\%$. 

\paragraph{Comparison with previous literature} The problem of a particle in a magnetic field was considered in both \citep{jin2022learning} and \citep{bajars2023locally}. The conservation law \eqref{Cons_law_momentum} has not been identified in these papers, thus we do not provide a direct comparison. The visual agreement of the solutions is just as good as the one demonstrated in \citep{jin2022learning,bajars2023locally}. The absolute error in solution and the Hamiltonian is also similar to that presented in \citep{bajars2023locally}. There are no Casimirs in the system, so there is no additional advantage over the methods presented in \citep{jin2022learning} or \citep{bajars2023locally}. 

Additionally, although the system \eqref{charged_particle_eq} is Poisson, it is not in the Lie-Poisson form. The fact that LPNets can solve this problem on par with other methods is a consequence of the particular form of the magnetic field $\mathbf{B}(\mathbf{x})$ and the fact that the corresponding integrals \eqref{gen_Phi} can be computed explicitly. Even though the system \eqref{charged_particle_eq} is not Lie-Poisson, we found it useful to present the results in order to show possible extensions of the methods of LPNets for more general systems. 

\subsection{Kirchhoff's equations for an underwater vehicle}\label{sec_Kirchhoff}

The motion of a neutrally buoyant underwater body is described by the Kirchhoff equations, see \citep{leonard1997stability,LeMa1997stability} for the discussion on the Lie-Poisson structure of these equations and the corresponding stability results. When the centre of gravity coincides with the centre of buoyancy, the system simplifies somewhat but still possesses a rich dynamics with several integrable and chaotic cases with a rich history of study \citep{holmes1998dynamics}. We will treat that particular system with coincidence of centres of buoyancy and gravity and show the applicability of LPNets approach there as well.

The Hamiltonian for Kirchhoff's equations consists of the kinetic energy of rotational motion  and the energy of the translational motion with velocity $\bv$ and mass tensor $\mathbb{M}$: 
\begin{equation} 
H (\bPi,\bp)= \frac{1}{2}  \bPi \cdot \mathbb{I}^{-1} \bPi   + \frac{1}{2} \bp \cdot \mathbb{M}^{-1} \bp\,.
\label{Underwater_body_Hamiltonian}
\end{equation} 
The Poisson bracket for the underwater vehicle is expressed as the Lie-Poisson bracket for the group of rotations and translations $SE(3)$: 
\begin{equation}
\{ F, H \} = - \bPi \cdot \left( \pp{F}{\bPi } \times \pp{H}{\bPi } \right) - \bp \cdot \left( \pp{F}{\bPi} \times \pp{H}{\bp} - \pp{H}{\bPi} \times \pp{F}{\bp} \right) \,.
\label{SE3_bracket}
\end{equation} 
This bracket has a specific form coming from the geometry of semidirect product groups, see \citep{holm1998euler,leonard1997stability,holm2009geometric}. 
Kirchhoff's equations of motion for the underwater vehicle are 
\begin{equation} 
\begin{aligned} 
\dot \bPi  & =  - \pp{H}{\bPi} \times \bPi - \pp{H}{\bp} \times \bp 
\\
\dot \bp & =  - \pp{H}{\bPi} \times \bp\,.
\end{aligned} 
\label{Underwater_body_eqs} 
\end{equation} 
Equations \eqref{Underwater_body_eqs} have two Casimirs: $C_1=\| \bp\|^2$ and $C_2=\bPi \cdot \bp$. In addition, the total energy given by \eqref{Underwater_body_Hamiltonian} is also conserved.

\paragraph{LPNets for Kirchhoff's equations} 
Bearing in mind that now we have two momenta $(\bPi, \bp)$, we shall take the following Hamiltonians: $H_1=\mathbf{A} \cdot \bPi$ and $H_2 =  \mathbf{b} \cdot \bp$. Equations of motion \eqref{Underwater_body_eqs} reduce to  
\begin{equation} 
\label{underwater_LPNets} 
\begin{aligned} 
& H_1=\mathbf{A} \cdot \bPi:  \quad \dot \bPi = -  \mathbf{A} \times \bPi\, , \quad \dot \bp = -  \mathbf{A} \times \bp\,  
\\ 
&H_2 = \mathbf{b} \cdot \bp :   \quad 
\dot \bPi = -  \mathbf{b} \times \bp \, , \quad \dot \bp = \mathbf{0} \,.
\end{aligned} 
\end{equation} 
The first motion is the simultaneous rotation of the vectors $\bPi$ and $\bp$ about the same axis $\mathbf{A}$, by the same amount, with a given angular velocity. This is the transformation $(\bPi, \bp) \rightarrow (\mathbb{R}( \mathbf{A}, \theta) \bPi,  \mathbb{R}( \mathbf{A}, \theta)  \bp)$, where $\mathbb{R}(\mathbf{A}, \theta) $ is the rotation matrix about the axis $\mathbf{A}$ by the angle $\theta$. The second motion creates the transformation $(\bPi, \bp) \rightarrow (\bPi - \mathbf{b} \times \bp, \bp)$. These transformation describe the coadjoint action of $SE(3)$ on an element $\mse(3)^*$, see \ref{app:appendix_geometric_LPNets}.

Based on \eqref{underwater_LPNets}, we thus suggest LPNets for $SE(3)$-based Lie-Poisson equations \eqref{Underwater_body_eqs}. 
\paragraph{Data preparation}
\begin{enumerate}
\item Select the training data couples $(\mathbf{y}^0,\mathbf{y}^f)$, where $\mathbf{y}:= (\bPi,\bp)$.
\item Find the rotation axis $\mathbf{A}_j$ and angles $\theta_j$ that takes $\bp_j^0$ to $\bp_j^f$, and the corresponding rotation matrix $\mathbb{R}(\mathbf{A}_j,\theta_j)$. This could be accomplished by finding rotation angles, for example, the Euler or Tait angles, introducing the rotation mapping $\mathbf{p}_j^0$ into $\mathbf{p}_j^f$. Simultaneous rotation about either $\mathbf{p}_j^0$ or $\mathbf{p}_j^f$ does not change the end result and must be discarded. 

While it is possible to proceed in this manner for higher-dimensional groups, for the three-dimensional groups we can utilize a shortcut based on the cross-product of the two vectors, as we have done in the case of a rigid body. 
We compute 
\begin{equation} 
\mathbf{A}_j = \frac{1}{h} \frac{\bp^0_j \times \bp_j^f}{\| \bp_j^0\| \| \bp_j^f\| }\,.
\label{A_calc_SE3}
\end{equation}
The vector $\mathbf{A}_j$ contains all the information necessary for the first step of the algorithm, namely simultaneous rotation. In order to reconstruct the vector $\mathbf{A}_j$ on each time step, we only need the components normal to the vector $\mathbf{p}_j^0$. These components can be found by defining two vectors spanning the plane normal to $\mathbf{p}_j^0$ in the following way.  Take a fixed vector, for example, $\mathbf{e}_1 = (1,0,0)^T$, and for each $\mathbf{p}^0_j$ define 
\begin{equation} 
\bxi^1_j =\frac{\bp^0_j \times \mathbf{e}_1}{\| \bp^0_j \times \mathbf{e}_1\| } \, , \quad 
\bxi^2_j  = \frac{\bp_j^f \times \bxi_1}{\|\bp^0_j \times \mathbf{e}_1 \| }\,.
\label{xi1_xi2_def}
\end{equation}
Since $(\bxi_1,\bxi_2)$ are unit length and orthogonal to each other and also to $\mathbf{p}_j^0$, the vector $\mathbf{A}_j$ defined by \eqref{A_calc_SE3} can be uniquely reconstructed by 
\begin{equation} 
\mathbf{A}_j = \sum_{a=1}^2 a^a_j \bxi^a_j\, ,  \quad 
a^a_j := \mathbf{A}_j \cdot \bxi^a_j \,.
\label{A_reconstruction}
\end{equation}
Variables $a^{1,2}_j$ are the first part of the input for any pair of data points. 
\item Find the vector $\mathbf{b}_j$ (or, more precisely, two of its components normal to $\bp_f$), such that $\bPi_j^f - \big(\mathbb{R}(\mathbf{A}_j, \theta _j\big)\bPi_0 - \mathbf{b}_j \times \bp_j^f)$ vanishes.
This is accomplished by the following calculation. Define two vectors $\mathbf{E}_{1,2}$ as 
\begin{equation} 
\mathbf{E}_1 =\frac{\bp^0_j \times \bp_j^f}{\| \bp^0_j \times \bp_j^f\| } \, , \quad 
\mathbf{E}_2 = \frac{\mathbf{E}_1\times \bp_j^f}{\|\mathbf{E}_1\times \bp_j^f \| }\,.
\label{E1_E2_def}
\end{equation}
Clearly, $\mathbf{E}_{1,2}$ are orthogonal to each other and also to the vector $\mathbf{p}_j^f$. Note that $\mathbf{E}_1$ is simply the normalized version of vector $\mathbf{A}_j$ given by \eqref{A_calc_SE3}. 
\item 
We only need to equalize the components of $\bPi^f_j-\bPi^0_j$  which are normal to $\bp_j^f$ on each time step. We thus define the coefficients $\widetilde{b}^{1,2}$ \footnote{We used tildes to distinguish the coefficients of expansion  $\widetilde{b}^{1,2}$ in the basis of moving vectors $ (\mathbf{E}_1,\mathbf{E}_2)$ from the coefficients $b^{1,2}$ in the fixed frame basis $(\mathbf{e}_1,\mathbf{e}_2,\mathbf{e}_3)$ }  according to 
\begin{equation}
\widetilde{b}^{1,2} = (\bPi^f-\bPi^0) \cdot \mathbf{E}_{1,2}\,, \quad 
\mathbf{b} = \widetilde{b}^{1} \mathbf{E}_1 +\widetilde{b}^{2} \mathbf{E}_2 
\label{tildeb_def}
\end{equation} 
One can see that by construction, using this algorithm exactly conserves the Casimirs $|\bp|^2$ and $\bPi \cdot \bp$ on every time step. 
\item Each pair of mappings $(\bPi^0_j,\bp^0_j) \rightarrow (\bPi^f_j,\bp^f_j)$ (six coordinates) is parameterized by four variables: $(a^{1,2}_j,\widetilde{b}^{1,2}_j)$. These four variables are the outputs of the neural networks, whereas $(\bPi^0_j,\bp^0_j)$ are the inputs. 
\end{enumerate} 

\paragraph{Neural network training}
Similar to the procedure for the rigid body, we generate $50$ trajectories starting in the neighborhood of $\bPi^0=(1,1,1)$ and $\bp^0 = (-1,1,2)$. The initial points for trajectories are randomly distributed in the phase space with a uniform distribution in a cube of the size $2 a =0.2$ in every direction. Each trajectory has a $1000$ points (not counting the initial point) separated by the time interval of $h=0.1$, with $50,000$ data points total used for learning. 

\paragraph{Evaluation} 
Evaluation of trajectories using Neural network mapping $(\bPi,\bp)$ to the variables $(\mathbf{A},\widetilde{b}^{1,2})$ proceeds in next steps: 
\begin{enumerate}
\item Suppose the initial conditions $(\bPi^0,\bp^0)$ are given. Using the Neural Network, estimate the values $(a^{1,2},\widetilde{b}^{1,2})$ corresponding to the inputs. 
\item For each starting point of the momentum $\mathbf{p}^0$ and the pair $a^{1,2}$ evaluated by the neural network, build two vectors $(\bxi_1,\bxi_2) \perp \mathbf{p}^0$ according to \eqref{xi1_xi2_def} and reconstruct the axis of rotation $\mathbf{A}$ according to \eqref{A_calc_SE3}. 

\item Compute the three-dimensional rotation matrix $\mathbb{R}(\mathbf{n}_\mathbf{A}, \phi)$ about the axis $\mathbf{n}_\mathbf{A} = \mathbf{A}/\|\mathbf{A}\|$ and the angle $\phi = \mbox{arc sin} \left( \| \mathbf{A} \| h/\| \bp_0\|^2\right) $. 
\item Compute the new linear momentum $\bp^f$ and intermediate angular momentum $\bPi^f$ according to 
\begin{equation}
    \bp^f = \mathbb{R}(\mathbf{n}_\mathbf{A}, \phi) \bp^0 \, , \quad 
    \bPi^* = \mathbb{R}(\mathbf{n}_\mathbf{A}, \phi) \bPi^0 \,.
    \label{Reconstruction_1}
\end{equation}
\item Compute the axes $\mathbf{E}_{1,2}$ according to \eqref{E1_E2_def} and update the angular momentum according to 
\begin{equation}
    \bPi_f = \bPi_* - \widetilde{b}_1 \mathbf{E}_1 - \widetilde{b}_2 \mathbf{E}_2\,. 
\end{equation}
\item Reset the final values $(\bPi^f,\bp^f)$ to the initial conditions and repeat Step 1. 
\end{enumerate}
One can see that the algorithm formulated above conserves the Casimirs $|\bp|^2$ and $\bPi \cdot \bp$ exactly (\emph{i.e.}, to machine precision) on every time step during the trajectory prediction phase. 

\paragraph{Simulation results} 
\text{  }\\

{\bf Data generation} We present the simulation results 
for the LPNets for the Kirchhoff's equations. The data were obtained using $500$ trajectories of $100$ time steps with the step $h=0.1$. All trajectories started in the neighborhood of $\bar \bPi^0 = (1,1,1)^T$  and $\bar \bp^0 = (-1,1,2)^T$. The initial conditions are uniformly distributed in a cube of the size $2 a = 0.2$ in every direction of the phase space. The mass matrix $\mathbb{M}$ and tensor of inertia $\mathbb{I}$ in \eqref{Underwater_body_eqs} are taken to be $\mathbb{M}={\rm diag}(1,2,1)$ and $\mathbb{I}={\rm diag}(1,2,1)$. 

{\bf Network parameters}
The $50,000$ data pairs are used to produce $10,000$ points $(\bPi_j,\mathbf{p}_j)$ as input and the parameters $(a^1,a^2,\tilde{b}^1,\tilde{b}^2)$ as outputs. A fully connected network with $6$ inputs, $4$ outputs and $6$ hidden layers with $64$ neurons each, is initialized to describe the mapping. The total number of trainable parameters for the network is 21,508. All activation functions are taken to be sigmoid. The network is trained using Adam algorithm with the step of $0.001$, using mean square error as the loss function. From the data, $80\%$ of data are used for training and $20\%$ for validation. The network is optimized using Adams algorithm with the learning rate of $0^{-3}$ decaying exponentially to $10^{-5}$ for $500,000$ epochs. After the optimization procedure, he loss and validation loss reaching the values of $1.65 \cdot 10^{-4}$ and   $1.35 \cdot 10^{-3}$, respectively. 

\begin{figure}
    \centering
    \includegraphics[width = 1 \textwidth]{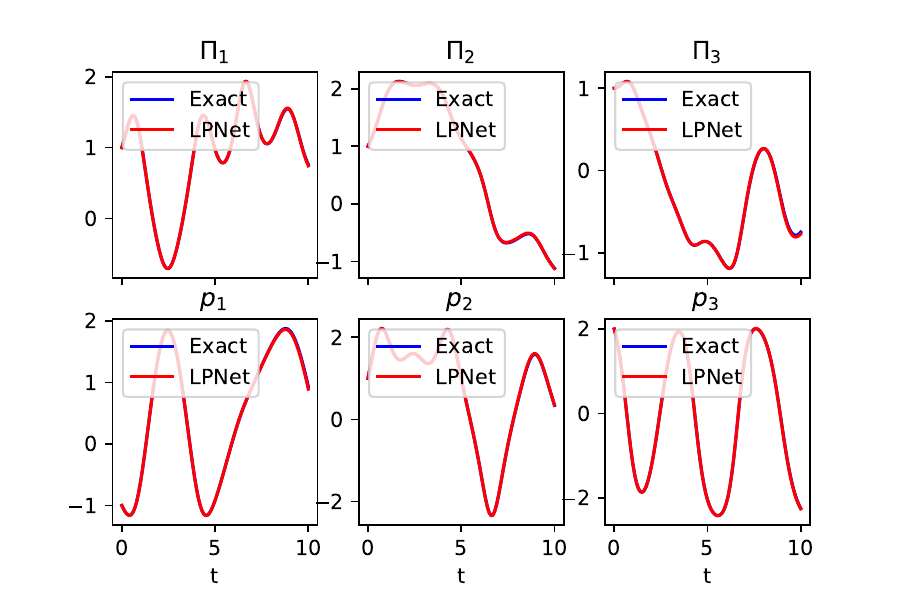}
    \caption{Results of simulation of equations \eqref{Underwater_body_eqs} and the corresponding solution of  using LPNets procedure. Upper panels: three components of the angular momenta $\bPi$; lower panels: three components of linear momenta $\bp$.  }
    \label{fig:SE3_Momenta}
\end{figure}

{\bf Evaluation of a trajectory}
A trajectory is reproduced using the LPNets algorithm as described above, using trained network, starting with the initial conditions $\bar \bPi^0 = (1,1,1)^T$  and $\bar \bp^0 = (-1,1,2)^T$ until $t=10$. As usual, the ground truth solution is produced by a high accuracy BDF algorithm with relative tolerance of $10^{-13}$ and absolute tolerance of $10^{-14}$. The comparison between the ground truth and the trajectories obtained by LPNets are presented on Figure~\ref{fig:SE3_Momenta}.

\begin{figure} 
\centering 
\includegraphics[width=0.48 \textwidth]{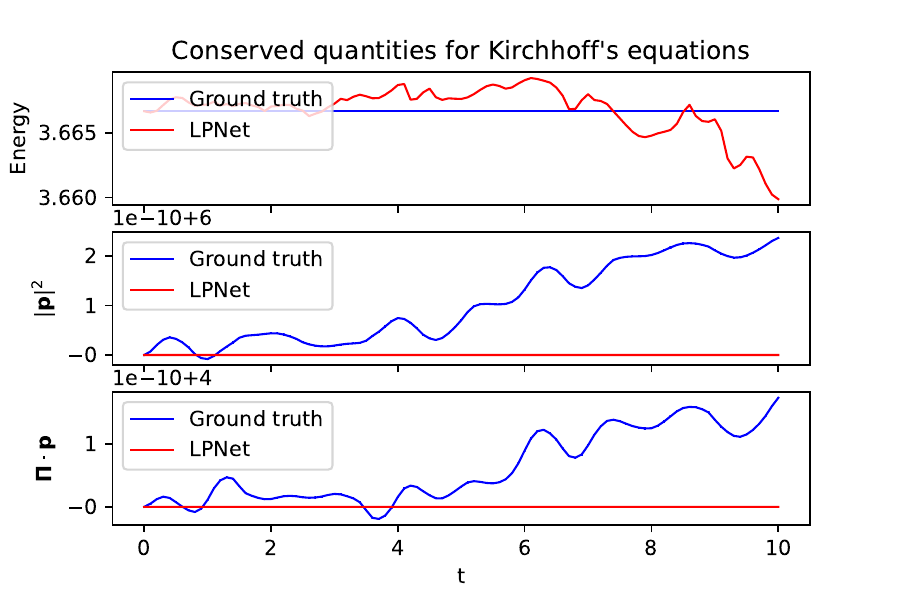}
\includegraphics[width=0.48 \textwidth]{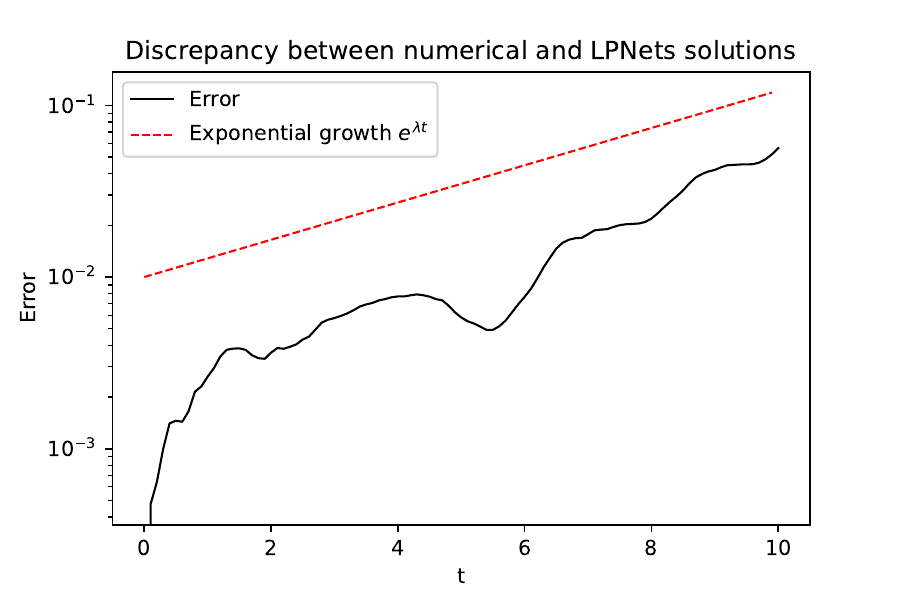}
\caption{Left: Conservation of the Hamiltonian $H$ (top) and two of the Casimirs $| \bp|^2$ (middle) and $\bPi \cdot \bp$ (bottom), comparing the results of LPNets (red) and ground truth (blue). The Hamiltonian is conserved to the relative accuracy of less than $0.3\%$. Notice that LPNets conserves the Casimir exactly (to machine precision) and thus exceeds the ground truth in the conservation of Casimirs. Right: The discrepancy between the results of LPNets and the ground truth. Dashed red line: growth of error expected from the Lyapunov exponent in the semilog scale. The growth of error follows the Lyapunov's exponent; thus, the neural network is performing as well as could be expected for a chaotic system.  \label{fig:LP_Conservation_error_SE3} }
\end{figure}

On Figure~\ref{fig:LP_Conservation_error_SE3}, left panel,  we show the conservation of the Hamiltonian (top), the first Casimir $| \bp|^2$ (middle) and the second $\bPi \cdot \bp$ (bottom). Notice that the Casimirs are conserved with higher accuracy than the ground truth, although the ground truth is already conserved to about $10^{-10}$. 

One may wonder whether better results can be obtained by using alternative ways of simulating trajectories or more effective implementations of neural networks.  
We have to keep in mind that Kirchhoff's equations \eqref{Underwater_body_eqs} are chaotic \citep{leonard1997stability,holmes1998dynamics}. We measure the rate of the divergence of nearby trajectories with the same values of the Casimirs, also known as the first Lyapunov exponent, to be about $\lambda = 0.250$ in units 1/time, starting with the same initial conditions as the simulated trajectory. The \emph{minimum} growth of errors expected by any numerical scheme is thus proportional to $e^{\lambda t}$ \citep{ott2002chaos}. On the right panel of Figure~\ref{fig:LP_Conservation_error_SE3} we present the semilogarithmic plot of the growth of errors, and show that it is  corresponds to the growth of errors expected from the chaotic properties of the system.  

\section{G-LPNets: Derivation and test cases}
\label{sec:G_LPNets}
\subsection{General theory}

Up until now, we have developed a method of computations using transformations that are easily computable, but the coefficients of these transformations must be obtained through the action of a Neural Network. We can extend these results and compute the transformation modules with parameters that are learned through an optimization procedure, producing modules preserving the Poisson structure of the bracket. We proceed as follows. 

For Lie-Poisson systems, Section~\ref{sec:Lie_Group_LPNets} showed that the flow generated by a test Hamiltonian, which is linear in momenta $H( \mu ) = \left< \alpha, \mu \right>$, can be computed analytically.  Instead of taking the test Hamiltonian to be a linear function of momenta, here we take the Hamiltonian to be $\widetilde{H}= \varphi(H)$, where $\varphi$ is some nonlinear scalar function of the variable $H$, which is also a constant of motion\footnote{Of course, a flow generated by  any  test Hamiltonian $H$ of an arbitrary functional form preserves $H$ and arbitrary function of that Hamiltonian $\varphi(H)$; but the flow generated by that Hamiltonian may not be explicitly solvable.}. Instead of equations \eqref{eq_LPNets}, we obtain: 
\begin{equation}
\dot \mu_a = \varphi'(H) C^d_{a b} \alpha^b \mu_d := \varphi'(H) \mathbb{M}(\alpha)^d_a  \mu_d =\varphi'(H) \mathbb{N}(\mu)_{a b}  \alpha^b
\,, \quad   H=  \left\langle \alpha , \mu _0 \right\rangle \, .
\label{eq_LPNets2}
\end{equation}
Noticing that $\varphi'(H)$ is constant under the flow, generated by the Hamiltonian $\widetilde{H} = \varphi(H)$, we conclude that the transformation generated by the flow of $\varphi(H)$ is simply obtained by a time scaling of \eqref{LPNets_transformation} and can be written as 
\begin{equation} 
\widetilde{\mathbb{T}}(\alpha,t) \mu_0 = 
\mathbb{T}(\alpha, \varphi'(H) t) \mu_0 = e^{\mathbb{M}(\alpha) \varphi'(H) t} \mu_0\,, \quad H=  \left\langle \alpha , \mu _0 \right\rangle \,,
\label{G_LPNets_transformation}
\end{equation} 
where $\mathbb{T}(\alpha,t)$ are defined as in \eqref{LPNets_transformation}. The prefactor $\varphi'(H)=\varphi '( \left\langle \alpha , \mu _0 \right\rangle )$ can be sought as a function of the variables using some kind of scalar activation function. Thus, 
in the G-LPNets framework, we are looking for transformations that are compositions of moduli $\mathbb{T}_s(a_s, \alpha_s,b_s, t)$, $s=1,\ldots M$ having the functional form  
\begin{equation} 
\widetilde{\mathbb{T}}_s(a_s, \alpha_s,b_s, t) \mu_0 
= e^{\mathbb{M}(\alpha_s) q_s t} \mu_0 \, , \quad 
q_s = a_s \sigma\left( \left\langle \alpha_s,\mu_0 \right\rangle \right) + b_s\,,
\label{G_LPNets_moduli}
\end{equation} 
where $\sigma$ is the activation function, which we will choose to be the sigmoid, and $(a_s, \alpha_s, b_s)$ are the parameters\footnote{ In the expression \eqref{G_LPNets_transformation}, $\alpha_s$ refers to the $s$ vector in the collection of $M$ vectors $(\alpha_1, \ldots, \alpha_M)$ in the Lie algebra $\mathfrak{g}$. In the example of the applications to the rigid body motion, for simplicity, we will put $\alpha_s$ to be proportional to the basis vectors of the Lie algebra with some scalar coefficients, which we will also call $\alpha_s$.}to be determined. 
The transformations \eqref{G_LPNets_transformation} are generalized moduli preserving the Poisson bracket and the Casimirs to machine precision. 

The method of G-LPNets selects the order of transformations $ \widetilde{\mathbb{T}}_s $ and minimizes a mean-square loss function. For example, for the data comprising $N$ points specifying the beginning $\mu_i^0$ and end $\mu_i^f$ values of momenta on each time interval, and assuming that the composition takes $M$ exactly equal time-substeps, the loss function can be taken to be the mean square error (MSE)
\begin{equation} 
L = \frac{1}{N} \sum_{i=1}^N \left\| \widetilde{\mathbb{T}}_M  (a_M, \alpha_M,b_M, h/M) \circ \ldots \circ \widetilde{\mathbb{T}}_M  (a_1, \alpha_1,b_1, h/M) \mu_i^0 -\mu_i^f \right\|^2 \, . 
\label{Loss_function_G_LPNets}
\end{equation}

The use of the mean square loss is advantageous since the derivatives of $L$ with respect to parameters $a_s, \alpha_s, b_s$ can be found analytically. Denoting the collection of parameters with a bar for shortness, for example, $\bar a = (a_1, a_2, \ldots) $ \emph{etc}, we naturally have 
\begin{equation} 
(\bar a, \bar \alpha, \bar b) = 
\mbox{arg min } L (\bar a, \bar \alpha, \bar b). 
\label{param_values}
\end{equation} 
There are several points to keep in mind regarding the application of G-LPNets. 
\begin{enumerate}
\item If there are no Casimirs, it is natural to choose the transformations alternating in a given sequence. For example, for an $n$-dimensional system one could choose a repeated application of $\widetilde{\mathbb{T}}_1$, $ \widetilde{\mathbb{T}}_2$,  $\widetilde{\mathbb{T}}_M$ followed by $\widetilde{\mathbb{T}}_1$ etc. It would be natural (although strictly speaking not necessary) to take the depth of network to be  $M=n \cdot k$, where $k$ is an integer number. 
\item If there are $d$ independent Casimirs $C_j(\mu)$, $j = 1, \ldots, d$, the transformations  producing the motion about the axis parallel to $\nabla C_j$ produce no effective evolution of momenta.
%\todo{ \color{red} $ \alpha _C$ strange notation?\\ 
%\textcolor{magenta}{VP: Indeed... Changed above. In fact, in the current notation, it is rather convoluted to write what the axis is. It is a highly nonlinear function of $(\bar a, \bar \alpha, \bar b)$ - that is why subtracting the motion about $\nabla C$ is not as easy here. It is indeed one of the disadvantages of G-LPNets (I think it is a disadvantage, at least).}}
One can either choose $n-d$ transformations on every time step by excluding certain $\alpha$, or simply apply all transformations in a sequence with the understanding that the resulting $\alpha$ is not unique. The latter is the approach we will use to describe rigid body rotation below. 
\item The advantage of applying all $\widetilde{\mathbb{T}}_s$ in a row is that there will be no accuracy loss if, at some point, the momentum is becoming close to parallel with the given coordinate axis. The disadvantage of applying all transformations in a sequence without excluding any $\alpha$ lies in the necessity for post-processing for \eqref{param_values}, as they are only defined up to corresponding components of gradients that are not in the span of the gradients of Casimir functions $\nabla_\mu C_j$, $j=1, \ldots, d$. In fact, even at the continuous level, the Hamiltonian is defined only up to the Casimirs, since the same dynamics are obtained if an arbitrary Casimir is added to the Hamiltonian. Similarly, all possible solutions of \eqref{param_values} obtained by G-LPNets are in the same equivalence class since they generate the same dynamics in phase space.
\item It is also possible that because parameters are defined only up to a certain vector or vectors, one could encounter vanishing gradients in some directions and a slow down of the convergence to the desired solution due to some numerical artifacts. We have not encountered these artifacts during our solution of the rigid body equations, but we cannot exclude further numerical difficulties when applying G-LPNets for general high-dimensional problems. 
\item  We expect that G-LPNets will also work for the cases beyond the Lie-Poisson framework, whenever explicit integration of the trajectories with the test Hamiltonians is possible, such as the particle in a magnetic field. The transformations $\widetilde{\mathbb{T}}_s$ are then computed as the generalizations of the Poisson transformations with the unknown time scaling coefficients. As long as the completeness of these transformations will be achieved, we expect them to provide efficient and accurate data-based computing methods for general Poisson problems. 
\end{enumerate}

The most challenging part of applying G-LPNets, in our opinion, is the lack of a general completeness result. It would be nice if the G-LPNets moduli \eqref{G_LPNets_moduli} satisfied a completeness result analogous to that of SympNets \citep{jin2020sympnets}. Right now, it seems unlikely that a general result may be proved valid for all Lie-Poisson brackets. Without specifying more information about the particular Lie group, progress in that area may be limited. However, the silver lining here is that for each particular problem, the symmetry of the problem is bound to be known \emph{a priori}, where the exact value of the Hamiltonian may or may not be known. Thus, if one focuses on particular problem at hand, a completeness result of transformations leading to G-LPNets  could be feasible to achieve. It is also possible that some of the methods of analysis performed in \cite{jin2020sympnets} for proving completeness of the modules in symplectic space would be applicable in the more general setting of a particular Lie-Poisson bracket.  

\subsection{Applications to rigid body dynamics}
To show the potential power of G-LPNets, we treat the equations of a rigid body, following up on our discussion in Sec.~\ref{sec:rigid_body}. Now, instead of learning from trajectories that stay close to the desired area, we aim to learn the whole dynamics and simply choose $50$ initial points uniformly distributed in a cube in momentum space $\bPi$, $-2 \leq \Pi_a \leq 2$, $a = 1,2,3$. Each trajectory is simulated with a high precision ODE solver and output is provided every $h=0.1$, providing $20$ data pairs, with the total of $1000$ data pairs. All  parameters of the system are exactly as in Sec.~\ref{sec:rigid_body}. 

Our comparison is the reconstruction of the dynamics of a rigid body which was already done in \citep{bajars2023locally}. Note, however, that \citep{bajars2023locally} takes all data pairs on \emph{the same} Casimir surface. In our opinion, since the Casimir surface depends on the initial conditions, a physical system, such as a satellite, could be observed on different Casimir surfaces due to the fact that the thrusters or other external forces have moved it from one Casimir surface to another outside the observation time. The Hamiltonian and physical parameters of the satellite are assumed to be the same, so it makes sense that the ground truth is generated with the same Hamiltonian, but different Casimirs. 

After data points are generated, a G-LPnet with $6$ transformations ($18$ parameters) is generated. 
%\todo{\color{magenta} Because of the scaling by $h/M$ that I put into the program (before, they were absorbed into the values of parameters), I could lower the number of parameters by a factor of 10 while keeping the accuracy! Only 6 (yes, six!!!) transformations (18 parameters) was enough to achieve incredibly accurate results.  I have no idea why the technique works so well. }
The transformations $\widetilde{\mathbb{T}}_i$ are rotations about the coordinate axes in the \emph{fixed} frame $\mathbf{e}_{1,2,3}$ by the angle $\phi_i$ (\emph{i.e.}, $e_1 = (1,0,0)^T$ \emph{etc.}) The rotations proceed in the sequence
\begin{equation} 
\widetilde{\mathbb{T}}_1 \rightarrow \widetilde{\mathbb{T}}_2 \rightarrow 
\widetilde{\mathbb{T}}_3 \rightarrow 
\widetilde{\mathbb{T}}_1 \rightarrow \ldots 
\label{Rot_sequence}
\end{equation}
At the given step $s$, the rotation angle $\phi_s$ for the given value of the momentum $\bPi^0$ is given by 
\begin{equation} 
\phi_s = a_s \sigma( \mathbf{A}_s \cdot  \bPi^0 ) + b_s \,.
\label{rot_angle_G_LPNets}
\end{equation}
The optimization finds the values of $(\bar \alpha,\bar \alpha, \bar b)$ minimizing the MSE loss function. The gradients of MSE functions with respect to parameters are computed analytically. We tried gradient descent methods and discovered that while they work satisfactorily, they do require quite a substantial number of epochs to converge, as was already observed in \citep{jin2022learning,bajars2023locally}. To make computation more efficient, we implemented an optimization procedure based on the Broyden–Fletcher–Goldfarb–Shanno (BFGS) algorithm \citep{fletcher2000practical} into SciPy: \url{www.scipy.org}. We let the BFGS algorithm run for $3500$ iterations achieving the value of the Loss of less than $5 \cdot 10^{-10}$, a procedure which is several orders of magnitude more efficient than the standard gradient descent-based algorithm. 

Using the values of parameters found by the optimization procedure, $10$ long solution with $10,000$ data points each with the time step of $h=0.1$ (max time $t_{max} = 1000$) were generated and compared with the ground truth solution obtained by a high precision ODE method. These long-term solutions were generated to compare with the results in \citep{bajars2023locally}.  The initial conditions $\bPi_0$ for these solutions were taken to be random numbers chosen from a random distribution from a cube $|\Pi_a| \leq 1.25$, $a = 1,2,3$. The value of the cube was somewhat smaller than the initial conditions so the solutions are guaranteed to remain within the area where the data were available. 

\begin{figure} 
\centering 
\includegraphics[height=0.27 \textheight]{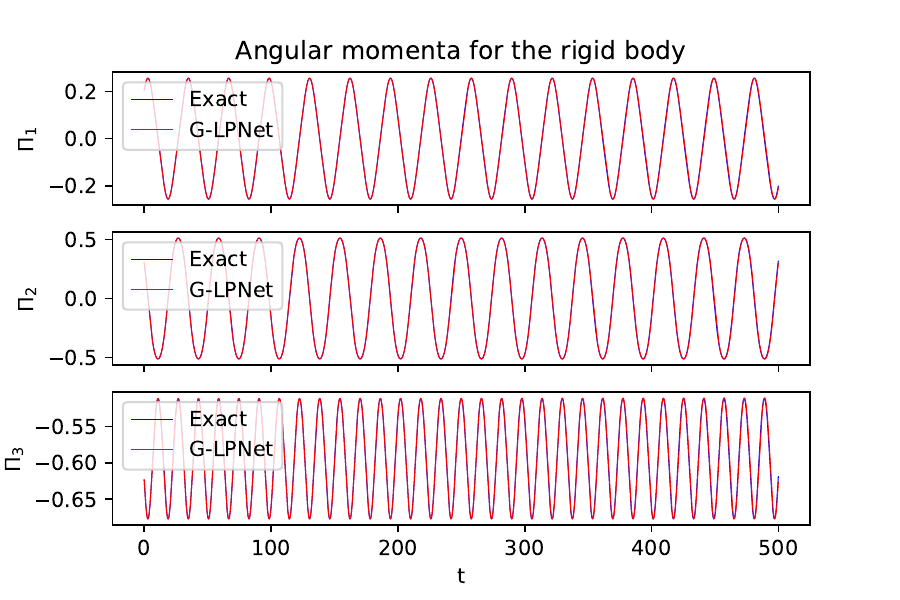}
\includegraphics[height=0.27 \textheight]{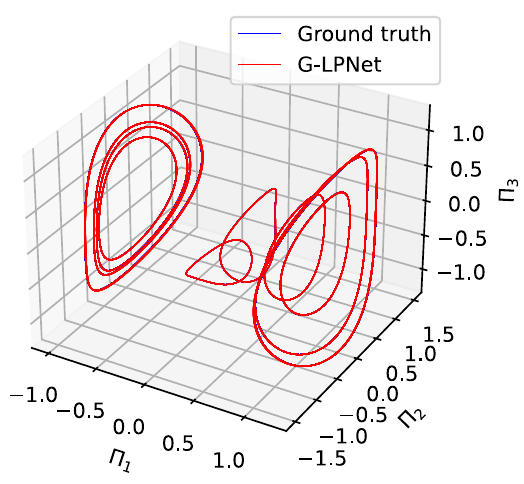}
\caption{Left: Results of G-LPNets applied to the motion of a rigid body (red) versus Ground truth (blue) for the individual momenta. Right: Parametric plot of the momenta in the phase space for $10$ solutions, chosen to start from uniformly distributed random points in the phase space in the cube $| \Pi_a | \leq 1.25$ in the phase space. The results of G-LPNets are indistinguishable from the ground truth solution.  \label{fig:G_LP_results_rigid_body} }
\end{figure}

\begin{figure} 
\centering 
\includegraphics[width=0.48 \textwidth]{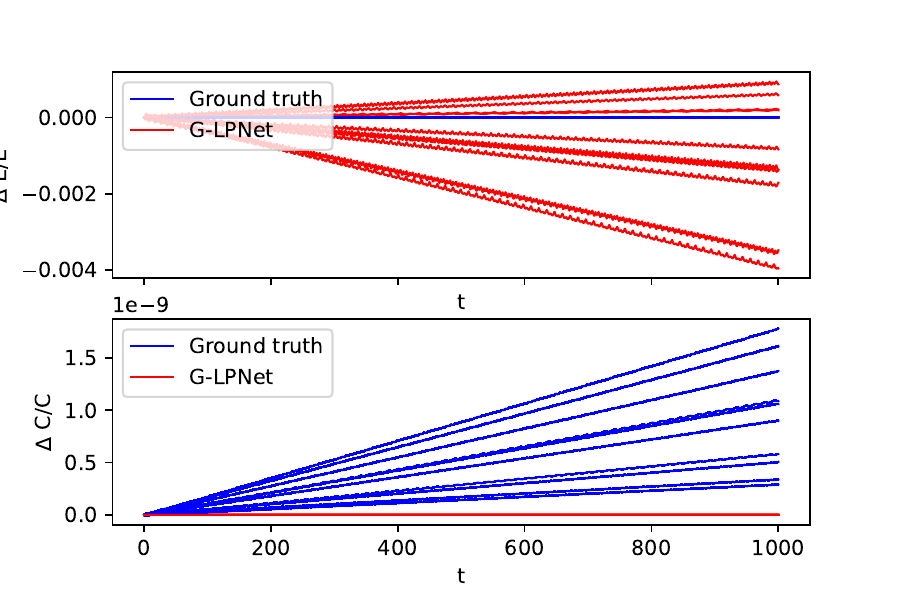}
\includegraphics[width=0.48 \textwidth]{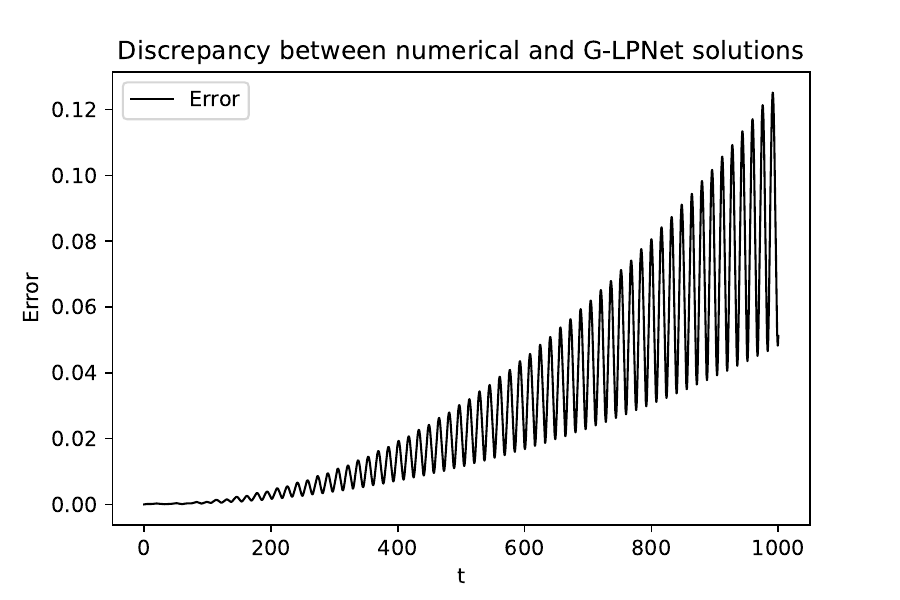}
\caption{Left panel, top: Relative accuracy for the conservation of the Hamiltonian $H$ (top) (\emph{i.e.}, $\Delta E/E = (E(t) - E(0))/\langle E\rangle$, where $\langle E\rangle$ is the mean value of the Hamiltonian (top). Left panel, bottom: the corresponding conservation of the Casimir $C$, computed as $\Delta C/C = (C(t) - C(0))/\langle C\rangle$ comparing the results of G-LPNets (red) and ground truth (blue), for all 10 simulations. As usual, G-LPNets conserve the Casimir exactly (to machine precision) and thus substantially exceed the ground truth in the conservation of Casimirs. Right: The discrepancy between the results of G-LPNets and the ground truth for the solution, presented on the left part of Figure~\ref{fig:G_LP_results_rigid_body}. Again, the discrepancy comes mostly from time mismatch, whereas the amplitude of oscillations is conserved with high precision due to the corresponding high precision in the conservation of energy. 
\label{fig:G_LP_Conservation_error_rigid_body} }
\end{figure}

In the left panel of Figure~\ref{fig:G_LP_results_rigid_body} we show a part of one of the $10$ trajectories extending to only about $t=500$ for clarity. This particular trajectory starts with the initial condition $\bPi_0 =(1.011,1.178,0.585)$, the results for other trajectories are similar. The values of momenta are nearly indistinguishable from ground truth for very long time. In the right panel of this Figure, we show all $10$ trajectories in the phase space, plotted for all times $0\leq t \leq 1000$ (10000 iterations). The results from the simulations and the ground truth coincide perfectly. This is due to the fact that both the Casimirs and the Hamiltonians are preserved with very high accuracy, as Figure~\ref{fig:G_LP_Conservation_error_rigid_body} shows. The Casimir is, again, conserved to machine precision on each step and thus substantially exceeds the accuracy of the calculation for the ground truth solution (still very high at about $10^{-9}$). The relative error in energy is of the order of $0.1\%$ over all long-term solution. 

\paragraph{Comparison with previous literature} The case of learning the whole dynamics of a rigid body was considered in \citep{bajars2023locally}. In that paper, the dynamics was considered only on a \emph{single} Casimir's surface $| \bPi | = 1$ with $300$ data points. The solutions generated by the neural network were also taken on that Casimir's surface. In contrast, we presented the dynamics with initial data taken from a volume of $\bPi$ (a cube), and initial conditions for G-LPNets are also taken without any restriction of the Casimir surface. Our relative error of solutions are of the same order as the results presented in \citep{bajars2023locally}. The conservation of Casimir and energy is reported to be of the order of $1.5\%$ in \citep{bajars2023locally}. In our case, the relative error in energy is somewhat better, with the max value of error around $0.4\%$. The Casimir $| \bPi|^2$ is conserved with the machine precision on every time step, so after $10^5$ steps a typical error in Casimir is expected to be of the order of $10^{-11} \div 10^{-10}$. While the relative error of energy in methods presented in \citep{bajars2023locally}  can no doubt be improved to reach the accuracy achieved here, we are not aware of any method preserving the value of the Casimir to the same precision as ours. 

\section{Conclusions}

We have derived a novel method of learning the evolution of a general Lie-Poisson system that is capable of predicting the dynamics in phase space with high precision. Our method learns the system by applying exact Poisson maps for certain test Hamiltonians, which are exactly solvable for any Lie-Poisson bracket. The resulting maps preserve the Poisson bracket under the evolution and also preserve all Casimirs with machine precision. These methods are also applicable to systems beyond Lie-Poisson, such as a particle in magnetic field, as long as the corresponding equations for test Hamiltonians are exactly solvable. 

We derive two types of networks. The first one is the Local Lie-Poisson Neural Networks (LPNets) which derive the local Poisson maps using test Hamiltonians that are linear in momenta. The parameters for these Poisson mappings resulting from the test Hamiltonians are then learned using standard methods of data-based learning with Artificial Neural Network (ANN). The advantage of this method is that the local completeness of the mappings is achieved automatically, since they represent exponential maps on a Lie group. An additional advantage of the method is the ability to use all the modern technology of ANNs for the discovery of the mapping from momenta to the parameters of test Hamiltonians. 

An alternative method was derived, called Global LPNets, using an example of nonlinear test Hamiltonians. These nonlinear Hamiltonians are arbitrary functions of the local Hamiltonians in the local LPNets approach. The explicit evolution maps on every time step obtained by these methods can be viewed as generalizations of the symplectic modules derived in \citep{jin2020sympnets} for case of a general Lie-Poisson bracket. We have presented an application of these methods to rigid body motion and showed that these methods demonstrate excellent accuracy and efficiency for long-term computation, and the ability to learn the dynamics in the whole phase space from quite a limited number of observations. 

While the G-LPNets seem more computationally effective than the local LPNets, we must caution the reader that a completeness result for mappings for G-LPNets obtained through the test nonlinear Hamiltonians is still missing. We believe it is probably highly unlikely that such a result can exist for a general Lie-Poisson system. The completeness result is likely to depend on the structure of the actual Lie group and the corresponding Lie-Poisson bracket. This is an interesting topic that we intend to address in the future. 

Another interesting topic is the question of the extension of this method to more general systems. In order to achieve that goal, the equations generated by the Poisson bracket for the test Hamiltonians must be exactly solvable, as it was in the case of the particle in the magnetic field. It will be interesting to compute the conditions on the Poisson bracket for such integrability to occur, which we also plan to undertake in the future.  Of particular interest are constrained systems, especially systems with nonholonomic constraints. The Lie-Poisson equations in this case become the Lie-Poisson-d'Alembert's equation, where the right-hand side of \eqref{Lie_Poisson_eq2} contains extra terms enforcing vanishing of momenta projections to certain subspaces defined by the constraints, such as in the Suslov problem \citep{Bl2003}. Apart from their importance in classical mechanics, these methods were recently found to be important for the variational discretizations of fluids  \citep{GaGB2020}, possibly including irreversible thermodynamics processes \citep{gawlik2022variational}. Extension of data-based computing to nonholonomic Lie-Poisson systems may thus play an important role in the applications of the method of this paper to continuum mechanics, including irreversible processes.
%%%END REM 

%From the two main approaches to continuum mechanics, the Lagrangian (coordinates tracking material particles) and the Eulerian (coordinates fixed in the spatial frame), the direct application to the Eulerian approach seems quite challenging. One can derive the equations for the test Hamiltonian and they will be solvable exactly by the method of characteristics. 

%A discretization of the continuum system will necessitate some kind of set of grid points from which characteristics would start. However, these characteristics would, as a rule, not land onto other grid points after the time step, and an interpolation to the grid points must be sought. We are not aware of any method that preserves the Poisson structure (or even symplectic structure) under interpolation. One could possibly circumvent these difficulties by adjusting the Poisson bracket slightly at each time step, or computing and compensating for the discrepancy in the Lie-Poisson evolution under interpolation. These interesting and important problems will be addressed in future work. 

\section{Acknowledgements}
We are grateful to Anthony Bloch, Pavel Bochev, Stephen Bond, Anthony Gruber, Melvin Leok, Tomoki Ohsawa, Tanya Schmah, Andrew Sinclair, Nathaniel Trask and Dmitry Zenkov for fruitful and engaging discussions. 
SH acknowledges support and experience provided by the internship in ATCO's transformation team and productive exchange with the team members. SH and VP were partially supported by the NSERC Discovery grant. 

This article has been co-authored by an employee of National Technology \& Engineering Solutions of Sandia, LLC under Contract No. DE-NA0003525 with the U.S. Department of Energy (DOE). The employee owns right, title and interest in and to the article and is responsible for its contents. The United States Government retains and the publisher, by accepting the article for publication, acknowledges that the United States Government retains a non-exclusive, paid-up, irrevocable, world-wide license to publish or reproduce the published form of this article or allow others to do so, for United States Government purposes. The DOE will provide public access to these results of federally sponsored research in accordance with the DOE Public Access Plan https://www.energy.gov/downloads/doe-public-access-plan.
%\todo{VP: Looks like \texttt{elsarticle-harv} with author-year option style fits what Neural Networks want \ldots Please go through the text and see that it looks OK. It doesn't give me any problems on compile like other styles. }

%\bibliographystyle{unsrt}
%\bibliographystyle{apalike}

\bibliography{References-LPNets}

\appendix

\section{Derivation of LPNets transformations in coordinate-free form}\label{app:appendix_geometric_LPNets}

We quickly review some facts on Lie-Poisson systems using intrinsic notations, and explain the emergence of the Lie-Poisson bracket as a consequence of the symmetry of the system.

\paragraph{Motion on Lie groups} Suppose $G$ is a Lie group and also the configuration manifold of the mechanical system. The equations of motion for the system with Lagrangian $L(g, \dot  g)$ follow as usual from Hamilton's principle. Assuming that the linear map $ \dot  g \mapsto p(g, \dot  g)= \frac{\partial L}{\partial \dot  g}( g, \dot  g)$ is invertible for all $g \in G$, one can equivalently formulate the dynamics via the \textit{canonical} Hamilton equations for the Hamiltonian $H(g,p)= \left\langle p, \dot g( g,p) \right\rangle - L(g, \dot  g(g,p))$, exactly as recalled in Sec.~\ref{subsec_intro} for vector spaces (except that care must be taken if one wants to formulate these equations in coordinate free form).

\paragraph{Lie group symmetries} Let us assume that the Lagrangian is invariant under the action of the group on itself by multiplication on the left: $L(hg, h\dot  g)= L(g, \dot  g)$, for all $h \in G$. One automatically gets that $H$ is also invariant: $H(hg, hp)= H(g, p)$, for all $h \in G$. It follows from this that $H$ is completely determined by it's dependence on momenta at the identity $g=e$, since we can write $H(g,p)= H(e, g ^{-1} p)=:h(g ^{-1} p)$. 
Here $ \mu = g ^{-1} p$ is the reduced (or body) momentum, which belongs the dual $ \mathfrak{g} ^* $ to the Lie algebra $ \mathfrak{g} $ of $G$, and $h: \mathfrak{g} ^* \rightarrow \mathbb{R} $ is the reduced Hamiltonian.
From this symmetry, it is naturally expected that the canonical Hamilton equations of motion for $(g(t),p(t))$ can be equivalently formulated by only using the reduced momenta $ \mu (t)= g(t) ^{-1} p(t)$, via an equation on $ \mathfrak{g} ^* $. This is indeed the case, which gives rise to the Lie-Poisson equations for $ \mu (t)$.

\paragraph{The Lie-Poisson equations} The Lie-Poisson equations in intrinsic form are written as
\begin{equation}\label{LP_intrinsic} 
\dot  \mu = \operatorname{ad}^*_{ \frac{\partial h}{\partial \mu } } \mu \,,
\end{equation} 
where $ \operatorname{ad}^*_ \xi : \mathfrak{g} ^* \rightarrow \mathfrak{g} ^*$ is the infinitesimal coadjoint action defined by $\langle\operatorname{ad}^*_ \xi \mu , \eta\rangle = \langle \mu , \operatorname{ad}_ \xi \eta \rangle$ with $\operatorname{ad}_ \xi \eta=[ \xi , \eta ]$, for $ \xi , \eta \in \mathfrak{g} $ and $ \mu \in \mathfrak{g} ^* $. This equation is the intrinsic version of \eqref{eqs_Lie_Poisson} written earlier. The evolution of a function $f( \mu )$ along the solution of \eqref{LP_intrinsic} is found as $ \dot  f= \{f,h\}$ with
\begin{equation} 
\left\{ f, h \right\} = - \left< \mu, \left[ \frac{\partial f}{\partial \mu } , \frac{\partial h}{\partial \mu } \right] \right> 
\label{Lie_Poisson_gen} 
\end{equation} 
the Lie-Poisson bracket. As we already commented earlier this Poisson bracket is noncanonical, however it is related to the canonical Poisson bracket $\{ \cdot , \cdot \}_{\rm can}$ governing the dynamics of $(g,p)$ as follows
\[
\left\{ f \circ \pi , h \circ \pi \right\}_{\rm can}= \left\{ f, h \right\} \circ \pi 
\]
with $ \pi (g,p)= g ^{-1} p= \mu $ the map sending the original momentum, to its reduced (or body) version. In other words, the map $ \pi $ is Poisson with respect to $\{ \cdot , \cdot \}_{\rm can}$ and $\{ \cdot , \cdot \}$, see Def. \ref{def_Poisson}. This natural process, a special instance of Poisson reduction, explains the occurrence of Lie-Poisson brackets in a large class of systems, see Table \ref{Table_Lie_Poisson_examples}. While some of these systems involve extensions of the process described above, the resulting noncanonical Poisson structures are each time justified as arising from a canonical Poisson bracket on a larger space (the initial phase space) by using the symmetries.

In the case of right-invariant systems, one has $ \mu = p g ^{-1} $ and a change of sign arises on the right hand side of \eqref{LP_intrinsic} and \eqref{Lie_Poisson_gen}. 

\paragraph{Flow of Lie-Poisson systems} Let us denote by $ \operatorname{Ad}_g: \mathfrak{g} \rightarrow \mathfrak{g}$ the adjoint action of $G$ on its Lie algebra, defined as $\operatorname{Ad}_g \xi := \left. \frac{d}{d\varepsilon}\right|_{\varepsilon=0} g c_ \varepsilon  g ^{-1}$ with $c_ \varepsilon \in G$ a curve tangent to $ \xi $ at $g=e$, and by $ \operatorname{Ad}^*_g: \mathfrak{g} ^* \rightarrow \mathfrak{g} ^*$ the coadjoint action defined as $ \langle \operatorname{Ad}^*_g \mu , \xi \rangle=\langle  \mu , \operatorname{Ad}_g\xi \rangle $. For $g(t) \in G$ we have the formula
\[
\frac{d}{dt} \operatorname{Ad}^*_{g(t)} \mu _0= \operatorname{ad}^*_{ g(t) ^{-1} \dot  g(t)} \operatorname{Ad}^*_{g(t)} \mu _0\,.   
\]
From this it immediately follows that the flow $ \phi _t( \mu _0)$ of the Lie-Poisson equations \eqref{LP_intrinsic} takes the form
\[
\phi _t( \mu _0)= \operatorname{Ad}^*_{g(t)} \mu _0 \quad\text{with}\quad  g(t) ^{-1} \dot  g(t)= \frac{\partial h}{\partial \mu }(t)
\]
and hence preserves the coadjoint orbits $ \mathcal{O} = \{ \operatorname{Ad}^*_g \mu _0\mid g \in G\} \subset \mathfrak{g} ^*$.

For the case of a linear Hamiltonian $h( \mu )= \langle a, \mu \rangle$, where $ \alpha \in \mathfrak{g}$ is a fixed element of the Lie algebra, the Lie-Poisson equation \eqref{LP_intrinsic} become
\begin{equation} 
\dot \mu =  \operatorname{ad}^*_ \alpha  \mu  
\label{Lie_Poisson_eq2} 
\end{equation} 
and its flow is found as
\[
\phi _t( \mu _0)= \operatorname{Ad}^*_{g(t)} \mu _0 \quad\text{with}\quad  g(t) ^{-1} \dot  g(t)= \alpha\,.
\]
This is the coordinate free version of \eqref{LPNets_transformation}, namely $\mathbb{T}(t, \alpha )$ is the coordinate version of $ \operatorname{Ad}^*_{ \operatorname{exp}( t \alpha )}$, with $ \operatorname{exp}$ the Lie group exponential. Then, we can derive LPNets if we can explicitly exponentiate the elements of the Lie algebra, so that it can be efficiently used in the minimization procedure \eqref{Loss_function}. 

For $SO(3)$ and $SE(3)$ the coadjoint action read
\[
\operatorname{Ad}_{\Lambda^{-1} } ^* \bPi = \Lambda \bPi  \quad\text{and}\quad \operatorname{Ad}_{(\Lambda, \mathbf{b})^{-1} } ^*( \bPi, \bp)= ( \Lambda \bPi + \mathbf{b} \times \Lambda \bp, \Lambda  \bp) 
\]
consistently with the formulas for $\mathbb{T}$ appearing in Sec.~\ref{sec:rigid_body}\&\ref{sec_Kirchhoff}.

%%%END REM 

\end{document}